\documentclass[10pt,journal,compsoc]{IEEEtran}

\ifCLASSOPTIONcompsoc
  \usepackage[nocompress]{cite}
\else
  \usepackage{cite}
\fi
\ifCLASSINFOpdf
   \usepackage[pdftex]{graphicx}
\else
   \usepackage[dvips]{graphicx}
\fi
\usepackage{cite}
\usepackage{amsmath,amssymb,amsfonts}
\usepackage{mathtools}
\usepackage{algorithmic}
\usepackage{subfig}
\usepackage[labelformat=parens,labelsep=quad,skip=3pt]{caption}
\usepackage{graphicx}
\usepackage{tabularx}
\usepackage{multirow}
\usepackage{array}
\usepackage{textcomp}
\usepackage{xcolor}
\usepackage{verbatim}
\usepackage{url}

\begin{document}

%
\title{Combination of Single and Multi-frame Image Super-resolution: An Analytical Perspective}
%
%
%
%

\author{Mohammad Mahdi~Afrasiabi,
Reshad~Hosseini
and~Aliazam~Abbasfar,~\IEEEmembership{Member,~IEEE,}
\IEEEcompsocitemizethanks{\IEEEcompsocthanksitem M. M. Afrasiabi, R. Hosseini, and A. Abbasfar are with the Department of Electrical and Computer Engineering, University of Tehran, Tehran, Iran. \protect\\ 
Email: {m.m.afrasiabi|rhosseini|abbasfar}@ut.ac.ir}
}

%
%

\markboth{Journal of \LaTeX\ Class Files,~Vol.~14, No.~8, March~2023}%
{Shell \MakeLowercase{\textit{et al.}}: Bare Demo of IEEEtran.cls for Computer Society Journals}
%



\IEEEtitleabstractindextext{%
\begin{abstract}
Super-resolution is the process of obtaining a high-resolution  image from one or more low-resolution images. Single image super-resolution (SISR) and multi-frame super-resolution (MFSR) methods have been evolved almost independently for years. A neglected study in this field is the theoretical analysis of finding the optimum combination of SISR and MFSR. To fill this gap, we propose a novel theoretical analysis based on the iterative shrinkage and thresholding algorithm. We implement and compare several approaches for combining SISR and MFSR, and simulation results support the finding of our theoretical analysis, both quantitatively and qualitatively.
\end{abstract}

}

\maketitle

\IEEEdisplaynontitleabstractindextext

%
\IEEEpeerreviewmaketitle

\IEEEraisesectionheading{\section{Introduction}\label{sec:introduction}}

%
%
%
%
\IEEEPARstart{I}{mproving} the resolution of images is one of the oldest research topics in image processing and machine vision. The process of transforming one or more low-resolution (LR) images to a high-resolution (HR) one is called super-resolution (SR) \cite{MFSR_Freq_1984}. Although the sensors (cameras) produced in recent years have relatively high quality and resolution, the need to increase the resolution of images is still substantial for the following reasons \cite{MFSR_Survey_2014, MFSR_Khattab_2018, MFSR_Milanfar_2019}. 1) Increasing the resolution of images recorded in the past with low-resolution cameras. 2) Increasing the resolution of medical or satellite images, which are usually expensive, and progress in their imaging technology is slow. 3) Combining the images taken by a handheld camera in a burst shot mode to achieve a higher quality. 4) Increasing the resolution of surveillance and security cameras, especially for facial recognition. 5) Saving the network bandwidth by sending LR images at the source and converting them to HR ones at the destination.
 
SR methods can be classified from different perspectives and in different ways  \cite{MFSR_Survey_2014, MFSR_Khattab_2018}. In a general and conventional division, there are two families of SR methods: SISR and MFSR. SISR works with only one input image while MFSR uses several input images to reach the final result. Although at the first glance, it may seem that SISR methods are special cases of MFSR ones, in general, the approaches and tools of these methods have major differences\footnote{It is important to mention that by MFSR methods, we mean those methods that use shift-and-add \cite{MFSR_Farsiu_2004} approach to solve the SR problem, and have not used SISR methods in their algorithm. For the methods that use both MFSR and SISR approaches, we use a different naming.} \cite{MFSR_Khattab_2018}. 

The SR problem is inherently ill-posed. Especially in SISR, it needs image priors to remove the problem from the under-determined condition \cite{SISR_Yang_2008}. The initial SISR methods tried to increase the resolution of the LR image with the help of different interpolation techniques \cite{SISR_Interp_1978, SISR_Interp_2001, SISR_Interp_2006}. Since the interpolation operation has inherent smoothness, this approach was ineffective for reconstructing high-frequency regions and edges. Although these methods do not perform well, they are still used in many commercial applications due to their simplicity. Later, methods were developed that used large number of image patches (previously prepared) to learn efficient mappings between LR and HR images \cite{SISR_Example_Internal_2009, SISR_Example_Internal_2011, SISR_Example_2002, SISR_Example_2004, SISR_Example_2013}. These example-based methods had a much higher performance than the interpolation-based ones. A branch of example-based methods used the assumption of the sparsity of natural images and developed methods based on solving the sparse coding (SC) problem with the help of LR and HR dictionaries prepared in advance \cite{SISR_Yang_2008, SISR_Yang_2010}. These methods are based on the assumption that the coefficients of the bases obtained by solving the SC problem for the LR image are approximately equal to  the coefficients obtained for the HR one. This assumption holds for natural images under some mild conditions \cite{SISR_Yang_2008}. Further, with the expansion of deep learning methods in computer vision, the field of SR was not the exception \cite{SISR_SRCNN_2014, SISR_SRCNN_2016, SISR_VDSR_2016, SISR_Residual_2018, wang2020deep}, and the first SR method based on deep learning was widely welcomed by researchers \cite{SISR_SRCNN_2014}. The authors of \cite{SISR_SRCNN_2014} showed better performance obtained by their proposed neural network framework compared to the previous SR methods. They demonstrated that the proposed neural network is somehow equivalent to the SC process, but since all its parameters are trained carefully, it achieves higher performance.

MFSR was first introduced in \cite{MFSR_Freq_1984} using a frequency domain approach only for the translational motions. \cite{MFSR_Elad_2001} proposed a fast method in pixel domain for an invariant point spread function and under additive white Gaussian noise (AWGN) condition like models in \cite{MFSR_Milanfar_2001,MFSR_Milanfar_2001_2}. Farsiu et al. in \cite{MFSR_Farsiu_2004} described a new method to consider other types of noise such as salt-and-pepper. \cite{MFSR_Farsiu_2004} is a strong work in the field of MFSR because of proposing a robust algorithm with clear mathematical justifications. 

Regardless of a few pieces of research that use a combination of single-image and multi-frame SR (ComSR) techniques such as \cite{Combined_Doubly_2015,Combined_Doubly_2016,Combined_M4S1_2017,Combined_S4M1_2020} in recent years, SISR and MFSR methods have been evolved separately and independently.
To the best of our knowledge, before the suggested method in \cite{Combined_Doubly_2015}, SC-based methods were employed only in SISR. As proposed in \cite{Combined_Doubly_2015} and \cite{Combined_Doubly_2016}, the LR images are concatenated, and then the corresponding sparse code is found.  
In \cite{Combined_S4M1_2020}, SR process is divided into two phases. At first, each LR image is magnified and recovered using a SISR method, and then an MFSR method is applied over the magnified images. We call this approach single first multi last (SFML)  ComSR approach. In \cite{Combined_M4S1_2017}, the process is done in the inverse order of \cite{Combined_S4M1_2020} where an MFSR method combines the LR images first, and then a SISR method is applied on the resulting image. We call this multi first single last (MFSL) ComSR approach. 

To the best of our knowledge, researchers have not yet theoretically answered the question of ``what is the optimal way to combine SISR and MFSR methods?''. Should the SISR be applied to the input images first, then the MFSR or vice versa? In this article, with the help of mathematical analysis, we will show the optimal way to combine SISR and MFSR. The main contribution of this paper is that we present an analytical approach on how to combine SISR and MFSR. We show that MFSL ComSR is optimal. Based on this analytical result, we also propose a novel SR method which is robust to relatively high AWGN.
Our analytical result is validated by numerous experiments. Especially when the up-scaling factor ($r$) is a  multiplication of two smaller integers ($r=r_1\times r_2$), we show that the optimal method of combining the LR images is still achieved by MFSL ComSR, that is by using MFSR (with $r$ up-scale) firstly, then applying a SISR network.

\section{Related Work}
The related work is divided into three subsections, each for an approach of solving the SR problem, namely SISR, MFSR and ComSR. 

\subsection{Single image super-resolution}
SISR methods can be divided as follows.
 
1) Interpolation-based methods. Early SISR approaches simply use interpolation (bi-linear, bi-cubic, Lanczos) and try to reconstruct new pixels by weighted averaging of neighboring pixels \cite{SISR_Interp_1978, SISR_Interp_2001, SISR_Interp_2006}. It is obvious that these methods are simple but have intrinsic smoothness (low-pass filtering). Therefore, they are not able to recover high-frequency contents, especially at the edges. 
 
2) Example-based (patch-based) methods. These methods use the existing paired HR and LR training images to find a proper mapping between HR and LR patches. Some of them obtain this mapping using internal learning (only internal patches of down-scaled and original given image, as LR and HR patches, correspondingly) \cite{SISR_Example_Internal_2009, SISR_Example_Internal_2011} and the other from external learning (given databases of LR and HR patches) \cite{SISR_Example_2002, SISR_Example_2004, SISR_Example_2013}. 

3) Sparse coding-based methods. These methods use the assumption of the possible sparse representation of natural images \cite{SISR_Yang_2008, SISR_Yang_2010}.
This assumption comes from the inherent redundancy of natural images.
These methods need to train over-complete LR and HR dictionaries to solve an SC problem \cite{SISR_Yang_2010}.
Each patch of an LR image could be represented by a linear combination of a few columns of the LR dictionary. The coefficients of this linear combination are as same as those we need to generate an HR patch (from corresponding columns of the HR dictionary) \cite{SISR_Yang_2010}.
 
4) Deep learning-based methods. With the progress of deep learning approaches in a broad spectrum of computer vision applications \cite{Other_CNN_2012, Other_CNN_2014, Other_CNN_2014_2}, a category of SISR methods was developed using deep learning techniques.
The pioneering work of \cite{SISR_SRCNN_2014} applied a simple convolutional neural network (CNN) to solve the SISR problem and outperformed the classical methods. This approach was matured in \cite{SISR_SRCNN_2016} and then has been developed in various ways by using
new architectures \cite{SISR_Wang_2015, SISR_VDSR_2016, SISR_Residual_2018, ma2021structure, gao2022bayesian, saharia2021image, huang2021transitional} or proper loss functions \cite{SISR_Loss_2016, SISR_Loss_2017, SISR_Loss_2018, SISR_Loss_2020, wang2021joint}.
Some new methods use real-world images instead of synthetic images to obtain better results in real scenarios \cite{xu2020exploiting, son2021toward, wang2021joint, gao2022bayesian}. In \cite{gao2022bayesian}, statistics of image priors are injected into the model and a bayesian framework is investigated.

\subsection{Multi-frame super-resolution}
MFSR methods can be divided as follows.

1) Frequency domain-based methods. These methods transfer the problem into the frequency domain (wavelet, DCT, and etc.) \cite{MFSR_Freq_1984, MFSR_Freq_1999, MFSR_Freq_2000, MFSR_Freq_2006, MFSR_Freq_2009}.
They model the shifts between LR images in a simple translational model, thus avoiding high computational complexity. These methods suffer from their high sensitivity to complex shifts and model mismatch.

2) Iterative back propagation. These methods are inherently iterative. They guess an initial estimate of the HR image by a simple intuition like averaging of the LR images and then try to refine that using an iterative optimization process \cite{MFSR_IBP_1990, MFSR_IBP_1996, MFSR_IBP_2010, MFSR_IBP_2011}.  

3) Regularization-based methods. These methods add some regularization to the cost function of the MFSR problem. These regularization terms convey image priors and hence result in more robust HR images with detail-preserving features\cite{MFSR_Reg_2001, MFSR_Farsiu_2004, MFSR_Reg_2007, MFSR_Reg_2016}. In fact, this technique is equivalent to a maximum a posteriori (MAP) solution with a special noise probability density function \cite{MFSR_Survey_2014}.

4) Deep learning-based methods.
These methods in MFSR are not as developed as in SISR \cite{MFSR_Deep_2022_DeepBurst}. However, some noticeable architectures have been developed for specific applications. \cite{MFSR_Deep_2017_face} improves the resolution of human face images. In this method, special features are extracted from each of the LR images and finally, with the help of these features and the relative shifts of LR images, the final image is reconstructed. \cite{MFSR_Deep_2020_HighResNet}  developed a recursive fusion framework for combining satellite images. This method uses encoder-decoder architecture. The method of \cite{MFSR_Deep_2020_DeepSum}, known as DeepSum, uses a CNN architecture and benefits from spatial and temporal correlations between LR images. In this method, LR images are registered implicitly. In \cite{MFSR_Deep_2020_sat, MFSR_Deep_2020_sat2, MFSR_Deep_2022_TR-MISR}, deep learning-based methods have been developed and used in satellite and remote sensing applications. In \cite{MFSR_Deep_2022_DeepBurst}, for general applications (such as burst SR in handheld photography), a method was proposed that explicitly uses the information of shifts between LR images for their combination. The final fusion uses an attention-based architecture.

\subsection{Combined SISR and MFSR}
Some researchers have made efforts to combine SISR and MFSR approaches. In \cite{Combined_Doubly_2015}, a method is introduced that extends the SC framework to multi-frame. In this approach, instead of solving the SC problem for one input image, the problem is solved for several input images by first registering LR images and then solving a new SC problem. This approach was proceeded in \cite{Combined_Doubly_2016}, wherein two phases of registration and solving the SC problem are presented in the form of solving a single optimization problem. Among other approaches on ComSR, we can refer to \cite{Combined_M4S1_2017}, in which the input LR images are first converted into a higher resolution image by a conventional MFSR method and then, the result passes through a SISR network. On the other hand, method of \cite{Combined_S4M1_2020} passes the input LR images separately through the SISR network and then combines the results of the previous step with the help of a conventional MFSR method. As it is clear, this method has a higher computational complexity than \cite{Combined_M4S1_2017} or other existing methods. In \cite{Combined_Afrasiabi_2020}, an optimization problem based on input LR images and their corresponding semi-HR images (generated by a SISR method) was proposed. This optimization problem has a closed form solution, which is actually a weighted combination of LR and semi-HR images.

\section{Background}
In this section, we provide an overview of the methods
needed for our analysis explained in the next section. These methods are the MFSR method based on a shift-and-add approach, the SISR method based on SC, and finally, the iterative shrinkage and thresholding algorithm (ISTA).

\subsection{MFSR based on shift-and-add}
 In SR applications, image degradation model is almost always described as \cite{MFSR_Farsiu_2004}
\begin{equation}
	\label{mainModel} 
	\mathbf{y}_k = \mathbf{R}\mathbf{F}_{k}\mathbf{H} \mathbf{x} + \mathbf{w}_{k},
\end{equation}
where $ \mathbf{x} \in \mathbb{R} ^{n} $ and $ \mathbf{y}_k \in \mathbb{R} ^{m} $  are HR image (ground truth) and $k$-th LR image (from $N$ LR images), respectively. For the sake of simplicity, all images are exhibited by vectors rather than matrices and their columns are stacked lexicographically. 
Also, $\mathbf{R} \in \mathbb{R} ^{m\times n}$, $\mathbf{F}_k \in \mathbb{R} ^{n\times n}$, and $\mathbf{H} \in \mathbb{R} ^{n\times n}$ are down-sampling, translation (shift), and  smoothing filter (blur) matrices (operators), respectively. AWGN is represented by vector $\mathbf{w}_k \in \mathbb{R} ^m$.

We define vector $ \mathbf{z}=\mathbf{Hx} $. Actually, the vector $ \mathbf{z}$ is the blurred version of the vector $ \mathbf{x}$.
A common way to obtain the HR image is defining a cost function in the form of $\mathbf{L}_2$ norm \cite{MFSR_Farsiu_2004,MFSR_Elad_2001} such as   

\begin{equation}
	\newcommand\norm[1]{\left\lVert#1\right\rVert}
	\rho(\mathbf{z}) = 
	\sum_{k=1}^N\norm{ \mathbf{RF}_{k}\mathbf{z-y}_{k} }^2.
\end{equation}
So, the optimum solution ($\widehat{\mathbf{z}}$) is given by
\begin{equation}
	\label{eq:z}
	\widehat{\mathbf{z}} = \sum_{k=1}^N {\mathbf{F}^T_k} {\mathbf{R}^T} {\mathbf{y}_k}.
\end{equation}

\subsection{SISR based on sparse coding}

Let $ \mathbf{x} \in \mathbb{R} ^{n} $ and $ \mathbf{y} \in \mathbb{R} ^{m} $ be HR and LR vectors (patches), respectively. Also, matrices $ \mathbf{D}_H \in \mathbb{R} ^{n\times K} $ and $ \mathbf{D}_L \in \mathbb{R} ^{m\times K} $ are HR and LR over-complete dictionaries, respectively. These dictionaries have been learned from thousands of training HR and LR pairs of patches \cite{SISR_Yang_2010}. SC-based methods state that $ \mathbf{y} $ can be represented as 
\begin{equation}
	\label{sparse-coding} 
	\mathbf{y} = \mathbf{D}_L \mathbf{\alpha},
\end{equation}
where $\mathbf{\alpha} \in \mathbb{R} ^{K}$ is a sparse vector. Then, under mild conditions, the vector $ \mathbf{x} $ could be reconstructed \cite{SISR_Yang_2010} using

\begin{equation}
	\widehat{\mathbf{x}} = \mathbf{D}_H \mathbf{\alpha}.
\end{equation}

\subsection{ISTA method}
ISTA method states that the solution of the SC problem, that is $\alpha$ in \eqref{sparse-coding}, could be obtained by the following fixed-point equation 
\cite{Other_ISTA_2004}
\begin{equation}
	\label{Lista}
	\alpha^{(n+1)}
	= h_{\tau} 
	\big\{ 
	{\mathbf{D}^T_L} {\mathbf{y}} + {\mathbf{S}} {\alpha^{(n)}}
	\big\},
\end{equation}
where $n$ and $h_{\tau}\{.\}$ indicate iteration index and shrinkage operator, and initial point is set to zero $\alpha^{(0)} = 0$. Matrix $\mathbf{S}$ is defined as
\begin{equation}
	\label{S}
	{\mathbf{S}} = L{\mathbf{I}} - 	{\mathbf{D}^T_L} {\mathbf{D}_L},
\end{equation}
where $\mathbf{I}$ and $L$ are identity matrix and upper bound of the largest eigenvalue of the matrix ${\mathbf{D}^T_L}{\mathbf{D}_L}$.

\section{Optimality Analysis of MFSL ComSR}
In this section, we show the optimality of MFSL ComSR method by deriving the ISTA method for a collection of LR images. To this end, we first derive a formulation of the ISTA method based on HR dictionary. Then, we show that the fixed-point equation of the ISTA method for the collection of LR images is the same as that of the single image ISTA method with a fused version of LR images as its input. This fused image can be considered as the result of a conventional MFSR method.
\subsection{ISTA formulation based on $\mathbf{D}_H$}
We represent the formulation of the ISTA method based on $\mathbf{D}_H$ rather than $\mathbf{D}_L$. We replace $\mathbf{D}_L$ in \eqref{Lista} by ${\mathbf{R}} {\mathbf{F}_0} {\mathbf{D}_H}$. In the case of SISR, translation matrix $\mathbf{F}_0$ is an identity matrix, hence it is not needed, but we intend to inject it in our formulation to use it the next subsection.
\begin{align}
\label{ISTA_new}
\alpha^{(n+1)}
&= h_{\tau} 
\big\{ 
{\mathbf{D}^T_L} {\mathbf{y}} + {\mathbf{S}} {\alpha^{(n)}}
\big\}
\nonumber \\ 
& = h_{\tau} 
\big\{ 
{\mathbf{D}^T_H} {\mathbf{F}^T_0} {\mathbf{R}^T} {\mathbf{y}} + L{\alpha^{(n)}} - {\mathbf{D}^T_H} {\mathbf{F}^T_0} {\mathbf{R}^T} {\mathbf{R}} {\mathbf{F}_0} {\mathbf{D}_H} {\alpha^{(n)}}
\big\}
\end{align}

\subsection{Optimality analysis}

We use all LR frames ($ \{\mathbf{y}_k\}_{k=1}^N $) effectively to estimate the HR image ($\mathbf{x}$). Similar to MFSR methods, we define vector $ \mathbf{z}=\mathbf{Hx} $ for simplifying the problem.
 We define a new vector $ \mathbf{y} $ including all the LR images as
\begin{equation}
	\label{y}
	\mathbf{y}^T=
	\begin{bmatrix}
		\mathbf{y}^T_{1}  & \mathbf{y}^T_{2} & \dots & \mathbf{y}^T_{N} \\
	\end{bmatrix}.
\end{equation}
Inspired by SC-based methods \cite{Combined_Doubly_2015}, we define $\mathbf{D}_L$, the corresponding LR dictionary of $ \mathbf{y} $ including all LR dictionaries of LR images as
\begin{equation}
	\label{Dl}	
	\mathbf{D}_L
	=
	\begin{bmatrix}
		\mathbf{D}_{1} \\
		\mathbf{D}_{2} \\
		\vdots \\
		\mathbf{D}_{N} 
	\end{bmatrix}
	=
	\begin{bmatrix}
		{\mathbf{R}} {\mathbf{F}_{1}} {\mathbf{D}_H} \\
		{\mathbf{R}} {\mathbf{F}_{2}} {\mathbf{D}_H} \\
		\vdots \\
		{\mathbf{R}} {\mathbf{F}_{N}} {\mathbf{D}_H} 
	\end{bmatrix}.
\end{equation}	
Inspired by the ISTA method, we try to solve the SC problem of \eqref{sparse-coding} using \eqref{y} and \eqref{Dl}. To this end, the matrix $\mathbf{S}$ of \eqref{S} needs to be computed. First, we calculate the matrix $ {\mathbf{D}^T_L} {\mathbf{D}_L} $  of \eqref{S},
\begin{align}
	{\mathbf{D}^T_L} {\mathbf{D}_L} 
	& = \sum_{k=1}^N {\mathbf{D}^T_k} {\mathbf{D}_k} 
	= \sum_{k=1}^N {{\mathbf{D}^T_H} {\mathbf{F}^T_{k}} {\mathbf{R}^T} {\mathbf{R}} {\mathbf{F}_{k}} {\mathbf{D}_H}} 
	\nonumber\\ 
	& = \sum_{k=1}^N {{\mathbf{D}^T_H} {\mathbf{C}_k} {\mathbf{D}_H}} 
	= {\mathbf{D}^T_H} {\Big[\sum_{k=1}^N {\mathbf{C}_k}\Big]} {\mathbf{D}_H} 
	\nonumber\\ 
	& = {\mathbf{D}^T_H} {\mathbf{C}} {\mathbf{D}_H},
\end{align}
where $	{\mathbf{C}_k} $ and $	{\mathbf{C}} $ are defined respectively by
\begin{equation}
	{\mathbf{C}_k} = {\mathbf{F}^T_{k}} {\mathbf{R}^T} {\mathbf{R}} {\mathbf{F}_{k}}, 
\end{equation}	
and
\begin{align}
	{\mathbf{C}} = \sum_{k=1}^N {\mathbf{C}_k}.
\end{align}
Therefore, the matrix $ \mathbf{S} $ of \eqref{S} is 
\begin{equation}
	\label{S2}
	{\mathbf{S}} 
	= L{\mathbf{I}} - {\mathbf{D}^T_L} {\mathbf{D}_L} 
	= L{\mathbf{I}} -  {\mathbf{D}^T_H} {\mathbf{C}} {\mathbf{D}_H}.
\end{equation}
Second, we calculate ${\mathbf{D}^T_L} {\mathbf{y}}$ of \eqref{Lista} as
\begin{equation}
	\label{Dty}
	{\mathbf{D}^T_L} {\mathbf{y}} 
	= \sum_{k=1}^N {\mathbf{D}^T_k} {\mathbf{y}_k}
	= \sum_{k=1}^N {{\mathbf{D}^T_H} {\mathbf{F}^T_{k}} {\mathbf{R}^T}} {\mathbf{y}_k} 
	= {\mathbf{D}^T_H} {\mathbf{y}_C},
\end{equation}
where fused image (shift-and-add) ${\mathbf{y}_C}$ is defined as
\begin{equation}
	\label{yc}
	{\mathbf{y}_C} = \sum_{k=1}^N { {\mathbf{F}^T_{k}} {\mathbf{R}^T}} {\mathbf{y}_k}.
\end{equation}
According to \eqref{Lista}, \eqref{S2}, and \eqref{Dty}, the fixed-point equation of ISTA is 
\begin{align}
	\label{ISTA_multi}
	\alpha^{(n+1)}
	&= h_{\tau} 
	\big\{ 
	{\mathbf{D}^T_L} {\mathbf{y}} + {\mathbf{S}} {\alpha^{(n)}}
	\big\}
	\nonumber \\ 
	& = h_{\tau} 
	\big\{ 
	{\mathbf{D}^T_H} {\mathbf{y}_C} + L{\alpha^{(n)}} - 	{\mathbf{D}^T_H} {\mathbf{C}} {\mathbf{D}_H} {\alpha^{(n)}} 
	\big\}.
\end{align}
The above fixed-point equation is very similar to single image ISTA fixed-point equation of \eqref{ISTA_new} with the difference that it has the fused image $\mathbf{y}_C$ as its input and the matrix $\mathbf{C}$ in \eqref{ISTA_multi} is the generalization of ${\mathbf{F}^T_{0}} {\mathbf{R}^T} {\mathbf{R}} {\mathbf{F}_{0}} $ in \eqref{ISTA_new}. Therefore, we can conclude that MFSL ComSR method is optimal. In the following subsection, we explain more on this.
\subsection{Discussion on optimality analysis}
We divide the discussion on optimality analysis into two parts. First, we consider the case where the up-scaling factor is an indivisible integer such as 3. Then, we further discuss the case for divisible integers such as 4 in the next part.
\subsubsection{Up-scaling factor is an indivisible integer}

\begin{figure}[t]
	\centering
	\vspace{-1.5 in}
	\includegraphics[width=3.7in,trim=4 4 4 4,clip]
	{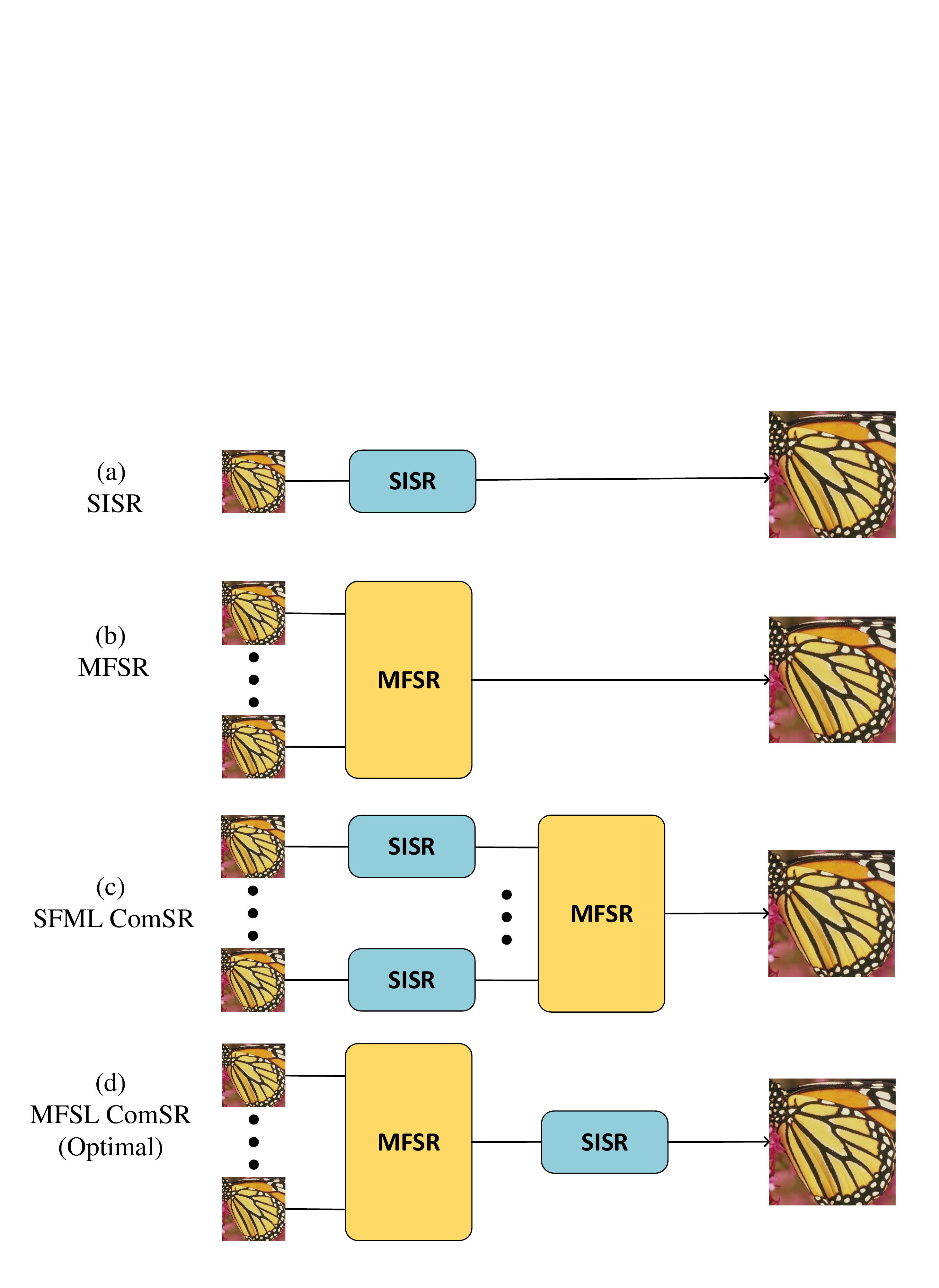}
	\caption{Various structures for combining SISR and MFSR (Up-scaling factor is an indivisible integer such as 3.)}
	\label{structures1}
\end{figure}

\begin{figure}[t]
	\centering
	\vspace{-2 in}
	\includegraphics[width=5.05in,trim=0 0 0 0,clip]
	{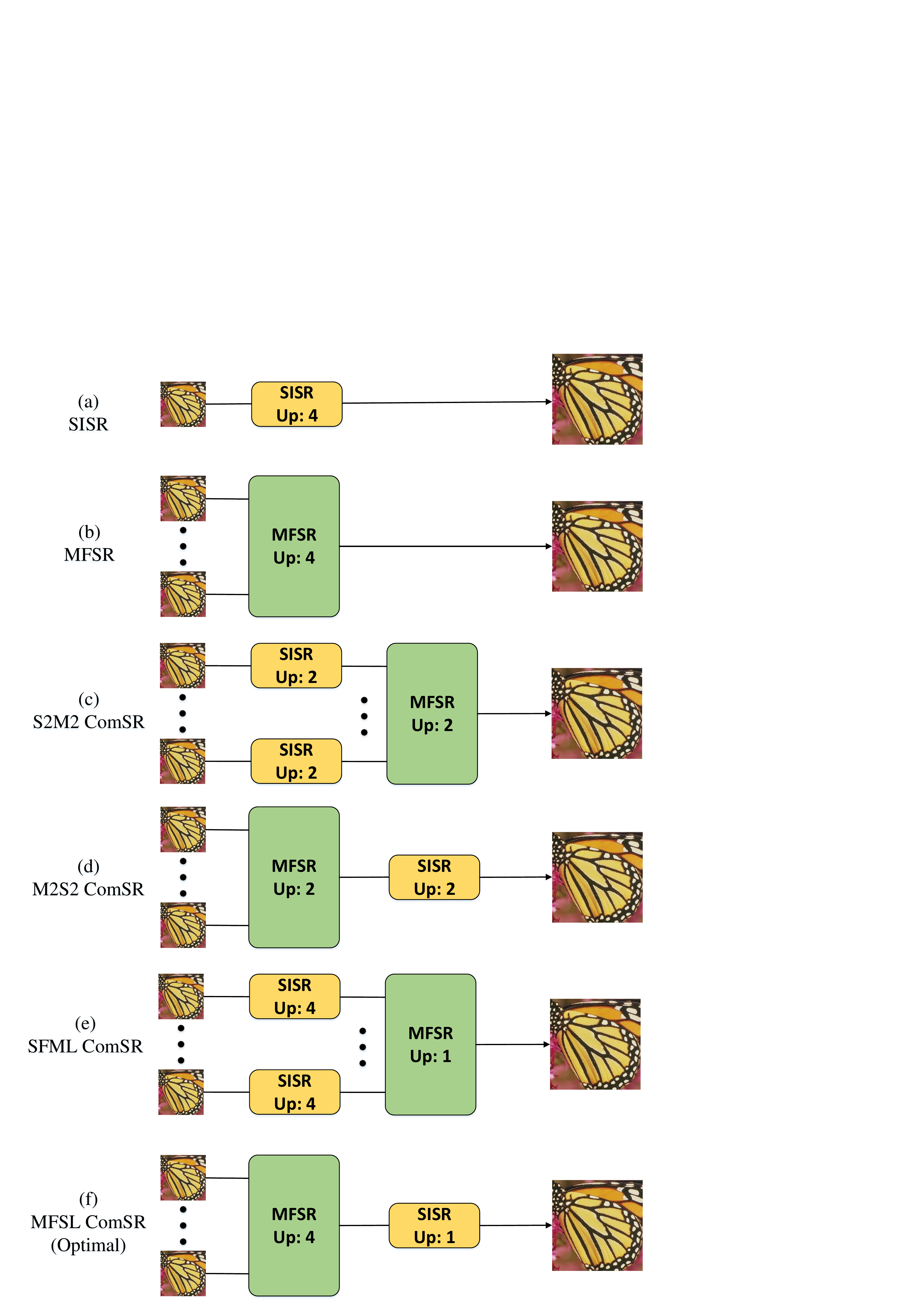}
	\caption{Various structures for combining SISR and MFSR (Up-scaling factor is a divisible integer such as 4.)}
	\label{structures2}
\end{figure}

Assuming the up-scaling factor to be an indivisible integer such as 2 or 3, one can consider either of four structures of  Fig.~\ref{structures1} to do the SR. 
To the best of our knowledge, the question of what is the best way (structure or method) to combine SISR and MFSR (when the up-scaling factor is an indivisible integer) has not yet been theoretically answered. Based on the examination of \eqref{ISTA_multi}, we understand that ${\mathbf{y}_C}$ should be built first, and then SISR should be used to improve it.
Considering \eqref{ISTA_new} and \eqref{ISTA_multi}, we realize that: 
\begin{itemize}
    \item[-]   \eqref{ISTA_new} is a special case of \eqref{ISTA_multi}.
\item[-] It can be seen from \eqref{ISTA_multi} that it first combines the up-sampled and shifted versions of the LR images (by generating ${\mathbf{y}_C}$). Then, it finds the solution with the help of the fixed-point equation of the ISTA method.

\item[-] Since \eqref{ISTA_multi} deals with the optimal solution of the sparse problem (through the iterative fixed-point algorithm introduced in \cite{Other_ISTA_2004}), we can conclude that the optimal approach of SR using multiple LR images needs the initial fused image to be computed by the combination of up-sampled and shift-corrected LR images.
Therefore, we suggest using the MFSL ComSR method by applying MFSR first and SISR after. 
\end{itemize}

\subsubsection{Up-scaling factor is a divisible integer}
There are several structures for performing SR using SISR, MFSR, or combinations of them when the up-scaling factor is a divisible integer such as 4 as shown in Fig.~\ref{structures2}. These structures are: a) Just using SISR \cite{SISR_SRCNN_2014} with $r=4$. b) Just using MFSR \cite{MFSR_Farsiu_2004} with $r=4$. c) Applying SISR on each LR image with $r=2$ and then combining them by MFSR with $r=2$, we call it S2M2 ComSR \cite{Combined_S4M1_2020}. d) Combining the LR images using MFSR with $r=2$ and then applying SISR with $r=2$ to improve the result and remove the artifacts, we call it M2S2 ComSR \cite{Combined_M4S1_2017}.
e) Applying SISR on each LR image with $r=4$ and then combining them by MFSR with $r=1$, we call it SFML ComSR \cite{Combined_S4M1_2020}.
f) Combining the LR images using MFSR with $r=4$ and then applying SISR with $r=1$ (it does not up-sample the image but uses SRCNN network to enhance the quality of output), called MFSL ComSR \cite{Combined_M4S1_2017}.

Similar to the case of indivisible up-scaling factor, based on the examination of \eqref{ISTA_multi}, it can be concluded that ${\mathbf{y}_C}$ should be built first, regardless of whether $r$ is divisible or indivisible. After that, SISR should be used to improve the final quality and to remove artifacts.

\section{Experimental Results}
For the experiments, we implement two different versions of the MFSR ComSR method. One is the MFSR ComSR method based on shift-and-add  \eqref{yc} in the case of noisy images, and the other is the two-dimensional curve fitting method (on LR images) in the case of noise-free images. In both versions, a registration method of \cite{Other_Reg_1992} is used to estimate the shift between an LR image and a randomly chosen reference LR image.
In the noise-free mode, it is enough to fuse the LR images with two-dimensional curve fitting method by estimating the pixel values in the main HR grid from shifted versions of LR images.
In noisy mode, the shift-and-add approach is used, meaning that we register LR images based on the estimated shifts between them, and then we re-sample and add the resulting images.
 The reason for the separation of MFSR ComSR into two modes (noisy and noise-free) is that in the noise-free mode, the degradation model of \eqref{mainModel} does not include AWGN and the averaging in \eqref{yc} causes the edges and high frequency regions to be damaged. 

For the SISR component in all experimented methods, we use the SRCNN method of \cite{SISR_SRCNN_2014}, a powerful method with low computational complexity. The network used in our simulations is exactly the same of \cite{SISR_SRCNN_2014}, which has been trained for the SISR task. In other words, we do not retrain or fine-tune the network of \cite{SISR_SRCNN_2014} in the simulations. Each average PSNR point is calculated over 20 independent trials. In each trial, for a specific dataset and number of LR frames, the LR images are generated by applying random shifts to the HR image of chosen dataset, then after down-sampling, AWGN with specific amount of noise is added\footnote{We provide the source codes of our simulations (implementations of evaluated methods) at {
\url{https://github.com/MMAfrasiabi/ComSR}}}.

Before conducting our comprehensive simulations and experimental results, we first examine the effect of ideal and practical image registration on the results of various SR methods.
\subsection{Effect of image registration methods}
At first, we examine the effect of practical and ideal image registration on the results of SR.
We use the method of \cite{Other_Reg_1992} for practical registration.
As illustrated in Table.~\ref{registration}, for Set5 and in two modes of noise-free and low-noise, the average PSNR values have been calculated for various methods, up-scaling factors, and the number of LR images (N). In the methods that require registration, there are two numbers in each cell of the table, which respectively correspond to practical and ideal registration. From examining the trend of the PSNR values, it can be concluded that even if we consider the registration as practical, we only see an almost constant decrease in the PSNR in various up-scaling factors and the number of LR images. Therefore, to remove the effect of registration error and to observe purely the effect of different structures and methods on SR performance, we use ideal registration in the following simulations.

\begin{table*}[!t]
	\renewcommand{\arraystretch}{1.5}
	\caption{Examining the effect of ideal and practical image registration on SR methods. Average PSNRs (in dB) are represented on Set5 dataset for different methods. Various up-scaling factors and number of input LR images (N) are included. In items with two values, first and second ones indicate average PSNR for practical and ideal registration, respectively.}
	\label{registration}
	\centering
	\begin{center}
		\begin{tabular}{ |c|c|c|c|c|c|c|c|c|c|c| } 
			\hline
			\bfseries \multirow{3}{*}{Dataset}  & \bfseries \multirow{3}{*}{Scale} & \bfseries \multirow{3}{*}{N} & \bfseries \multirow{3}{*}{Bicubic} & \bfseries SISR & \bfseries MFSR & \bfseries SFML ComSR & \bfseries Curve-Fit & \bfseries MFSL ComSR & \bfseries MFSL ComSR\\
			& & & & \multirow{1}{*} {SRCNN \cite{SISR_SRCNN_2014}}  & \multirow{1}{*} {Farsiu \cite{MFSR_Farsiu_2004}}  & \multirow{1}{*} {Kawulok \cite{Combined_S4M1_2020}}  &  & Optimal & Optimal\\
			 & & & & & & & & (Curve-Fit) & (Shift-and-Add) \\
			\hline \hline
			
			\multirow{4}{*}{Set 5} & 
			\multirow{2}{*}{$ \times 2 $} & 
			2 & \multirow{2}{*}{33.67} & \multirow{2}{*}{36.69} & 33.41/33.42 & 36.13/36.13 &  33.45/33.50 & 36.28/36.31 & 36.23/36.24\\ 
			\cline{3-3} 
			\cline{6-10}
			& & 3 &   &  & 33.44/33.45 & 36.18/36.18 & 33.49/33.51 & 36.39/36.40 & 36.30/36.34 \\
			
			\cline{2-10}
			& \multirow{3}{*}{$ \times 3 $} & 
			3 & \multirow{3}{*}{30.39} & \multirow{3}{*}{32.75} & 30.07/30.07 & 32.44/32.47 & 30.11/30.12 & 32.58/32.59 & 32.51/32.52 \\ 
			\cline{3-3} 
			\cline{6-10}
			(Noise-Free) & & 5 & & & 30.09/30.10 & 32.52/32.52 & 30.14/30.14 & 32.66/32.66 & 32.56/32.57 \\
			\cline{3-3} 
			\cline{6-10}
			& & 7 &  &  & 30.10/30.10 & 32.53/32.58 & 30.16/30.16 & 32.70/32.71 & 32.58/32.58\\
			\cline{2-10}
			
			\hline
			
			\multirow{5}{*}{Set 5} & 
			\multirow{2}{*}{$ \times 2 $} & 
			2 & \multirow{2}{*}{32.11} & \multirow{2}{*}{32.97} & 32.60/32.62 & 33.63/33.88 & 32.39/32.42 & 33.55/33.64 & 34.09/34.17\\ 
			\cline{3-3} 
			\cline{6-10}
			& & 3 & & & 32.91/32.94 & 34.64/34.63 & 32.52/32.57 & 33.86/33.95 & 34.83/34.92\\
			\cline{2-10}

			& \multirow{3}{*}{$ \times 3 $} & 
			3 & \multirow{3}{*}{29.71} & \multirow{3}{*}{31.45} & 29.93/29.94 & 32.08/32.11 & 29.82/29.85 & 31.82/31.84 & 31.98/32.02 \\ 
			\cline{3-3} 
			\cline{6-10}
			($\sigma_{n}=0.001$) & & 5 & & & 29.93/30.00 & 32.22/32.29  & 29.83/29.89 & 31.95/31.97 & 32.22/32.29\\
			\cline{3-3} 
			\cline{6-10}
			& & 7 & & & 29.96/30.01 & 32.36/32.38 & 29.88/29.91 & 31.97/32.01 & 32.25/32.38\\
			\cline{2-10}
			
			\hline
		\end{tabular}
	\end{center}
\end{table*}

\subsection{Up-scaling factor is an indivisible integer}

\begin{figure*}[htbp]
   	\centering
   	\begin{tabular}{cc}
   		\vspace{-0 pt}	
   		\subfloat[Set5, Noise-Free]{\includegraphics[width = 90 mm, scale = 1]{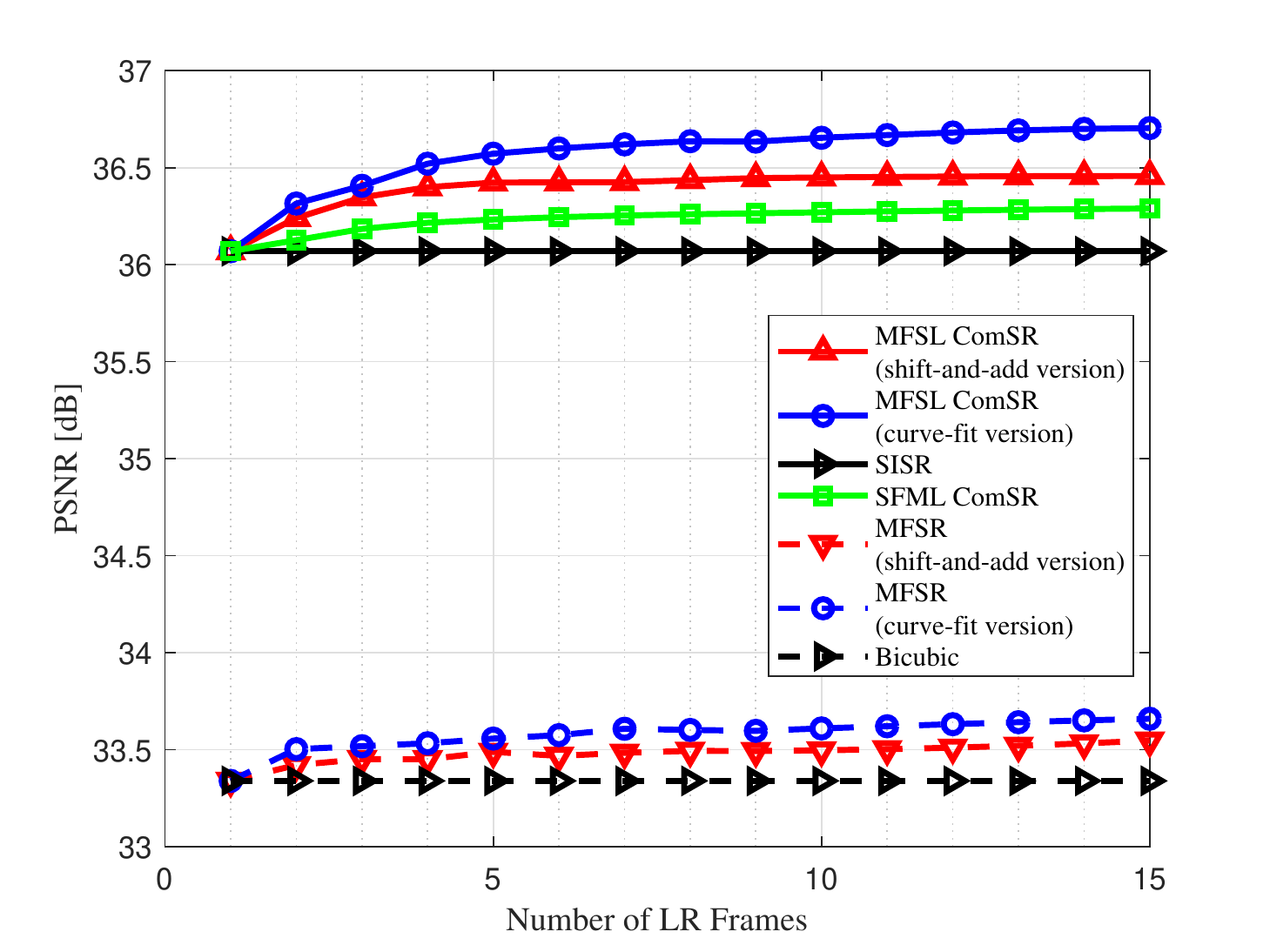}} & 
   		\subfloat[Set14, Noise-Free]{\includegraphics[width = 90 mm, scale = 1]{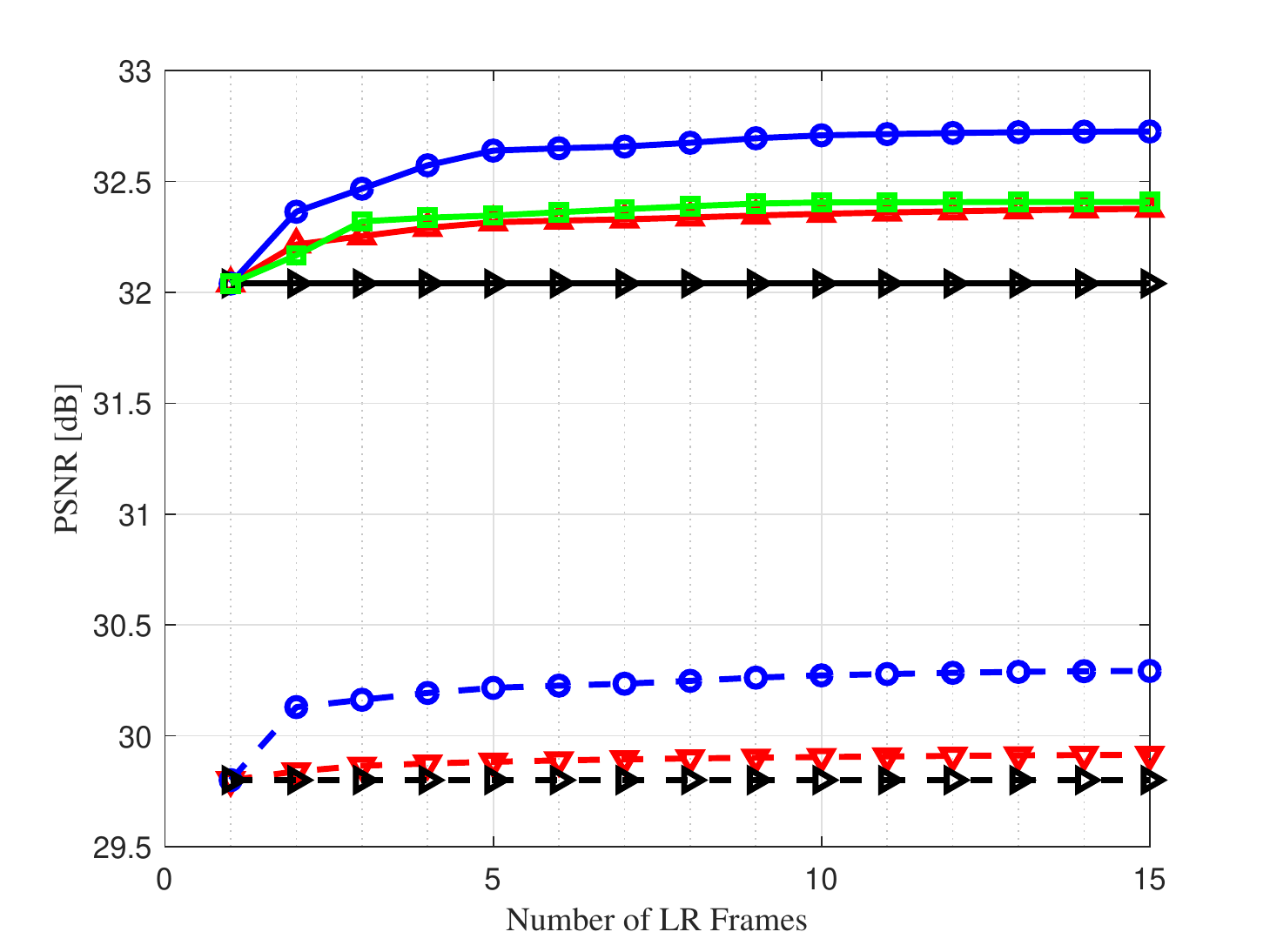}}  
   		\\
   		\subfloat[Set5, $\sigma_{n}=0.001$]{\includegraphics[width = 90 mm, scale = 1]{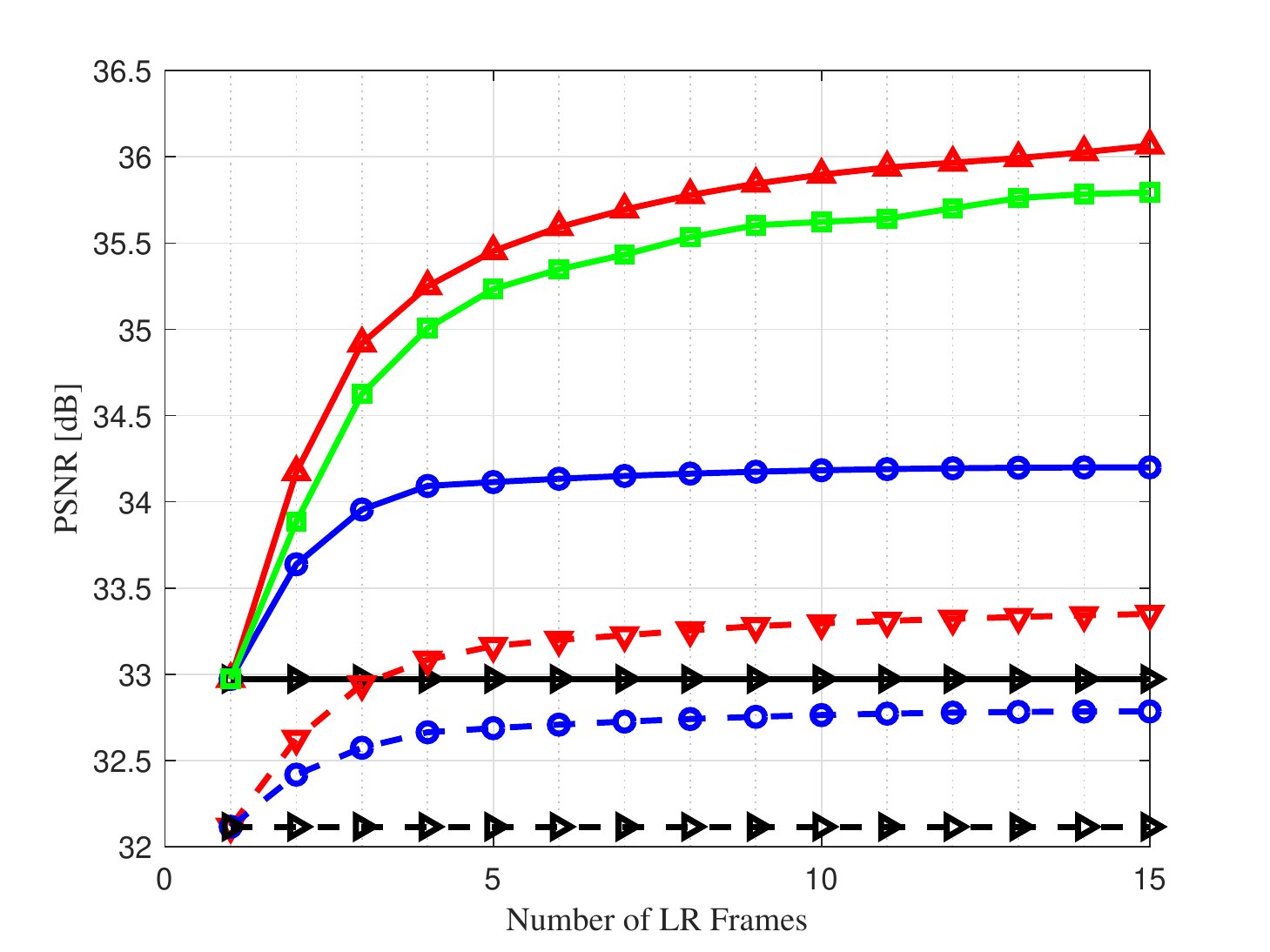}}  
   		& 
   		\subfloat[Set14, $\sigma_{n}=0.001$]{\includegraphics[width = 90 mm, scale = 1]{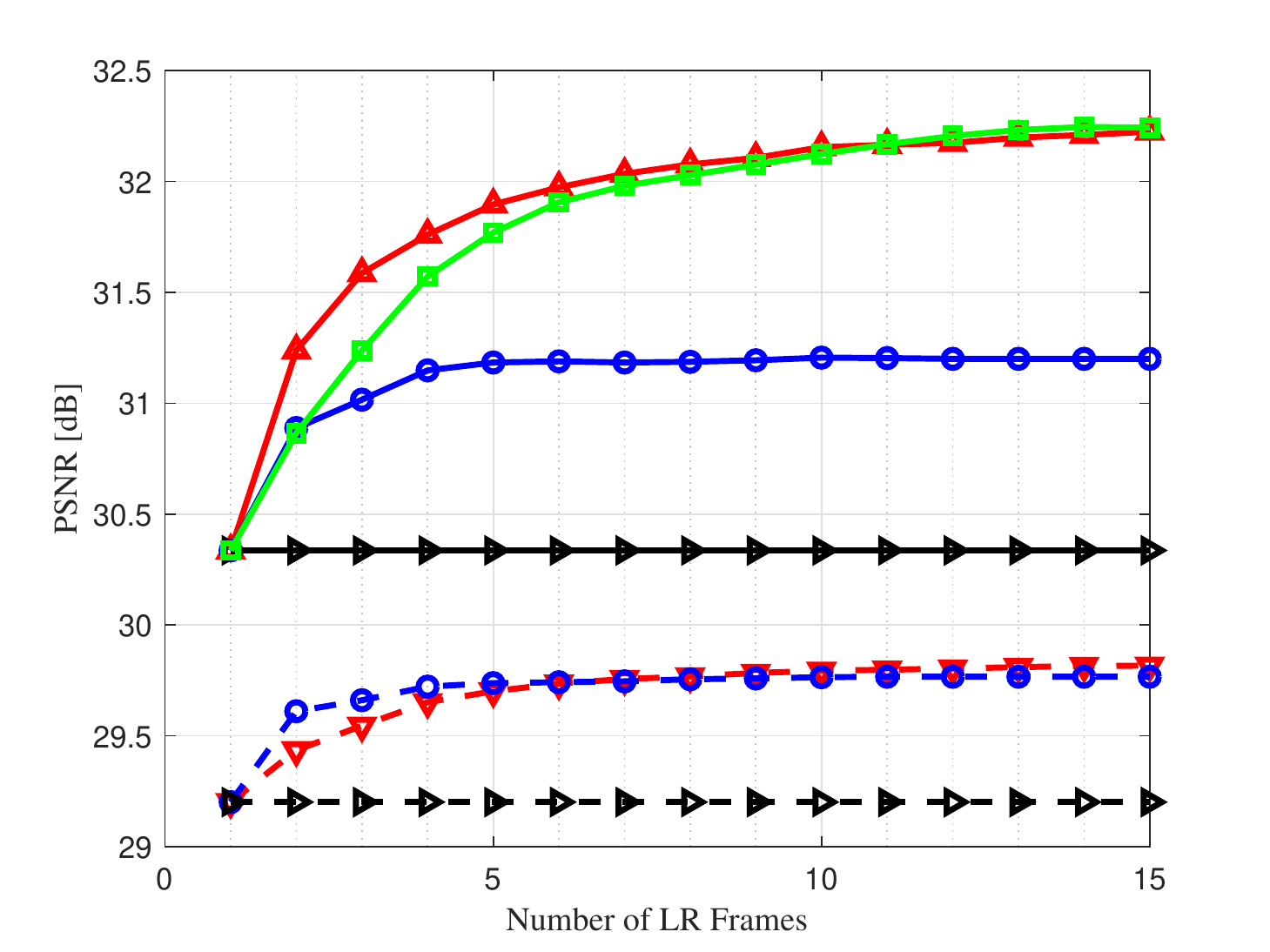}} 
   		\\
   		\subfloat[Set5, $\sigma_{n}=0.005$]{\includegraphics[width = 90 mm, scale = 1]{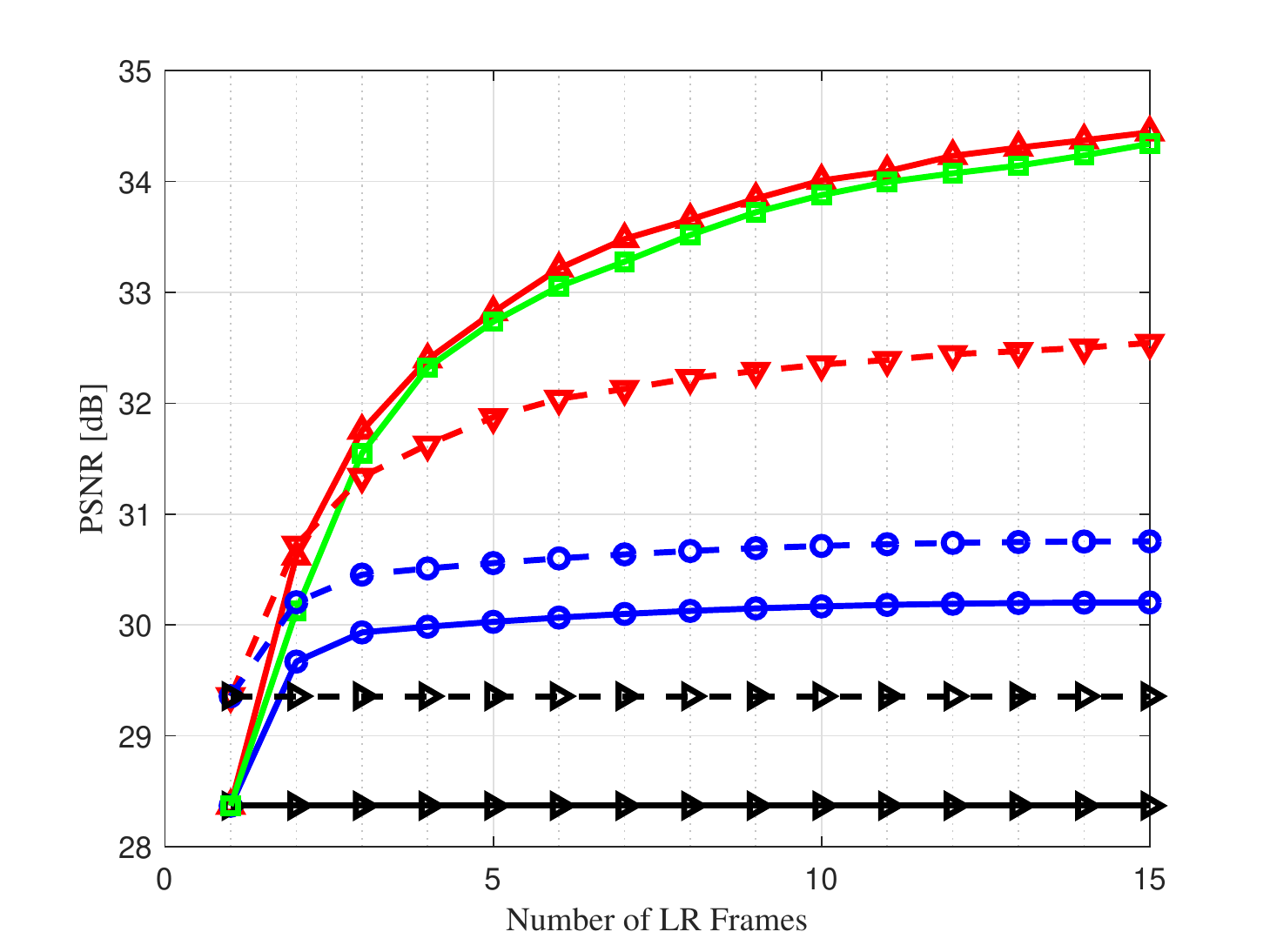}}  
   		& 
   		\subfloat[Set14, $\sigma_{n}=0.005$]{\includegraphics[width = 90 mm, scale = 1]{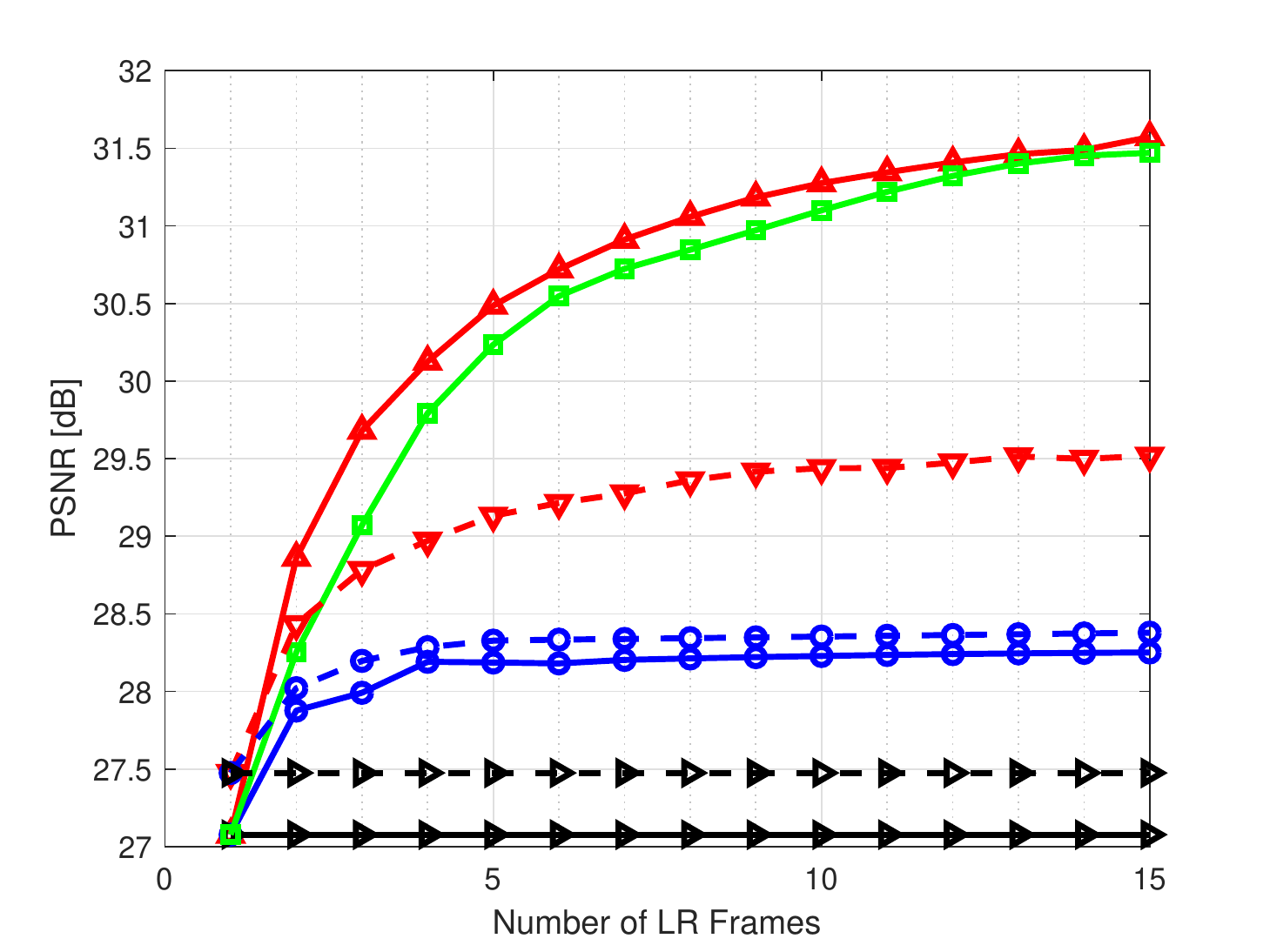}}
   	\end{tabular}
   	\caption{Average PSNR (in dB) vs number of LR frames for up-scaling factor $\times2$. Simulation results on Set5 dataset for noise-free, $\sigma_{n}=0.001$, and $\sigma_{n}=0.005$ are illustrated in (a), (c), and (e), respectively.
    Simulation results on Set14 dataset for noise-free, $\sigma_{n}=0.001$, and $\sigma_{n}=0.005$ are illustrated in (b), (d), and (f), respectively. }
   	
   	\label{sim_x2}
   \end{figure*}
  
\begin{figure*}[htbp]
     	\centering
     	\begin{tabular}{cc}
     		\vspace{-0 pt}	
     		\subfloat[Set5, Noise-Free]{\includegraphics[width = 90 mm, scale = 1]{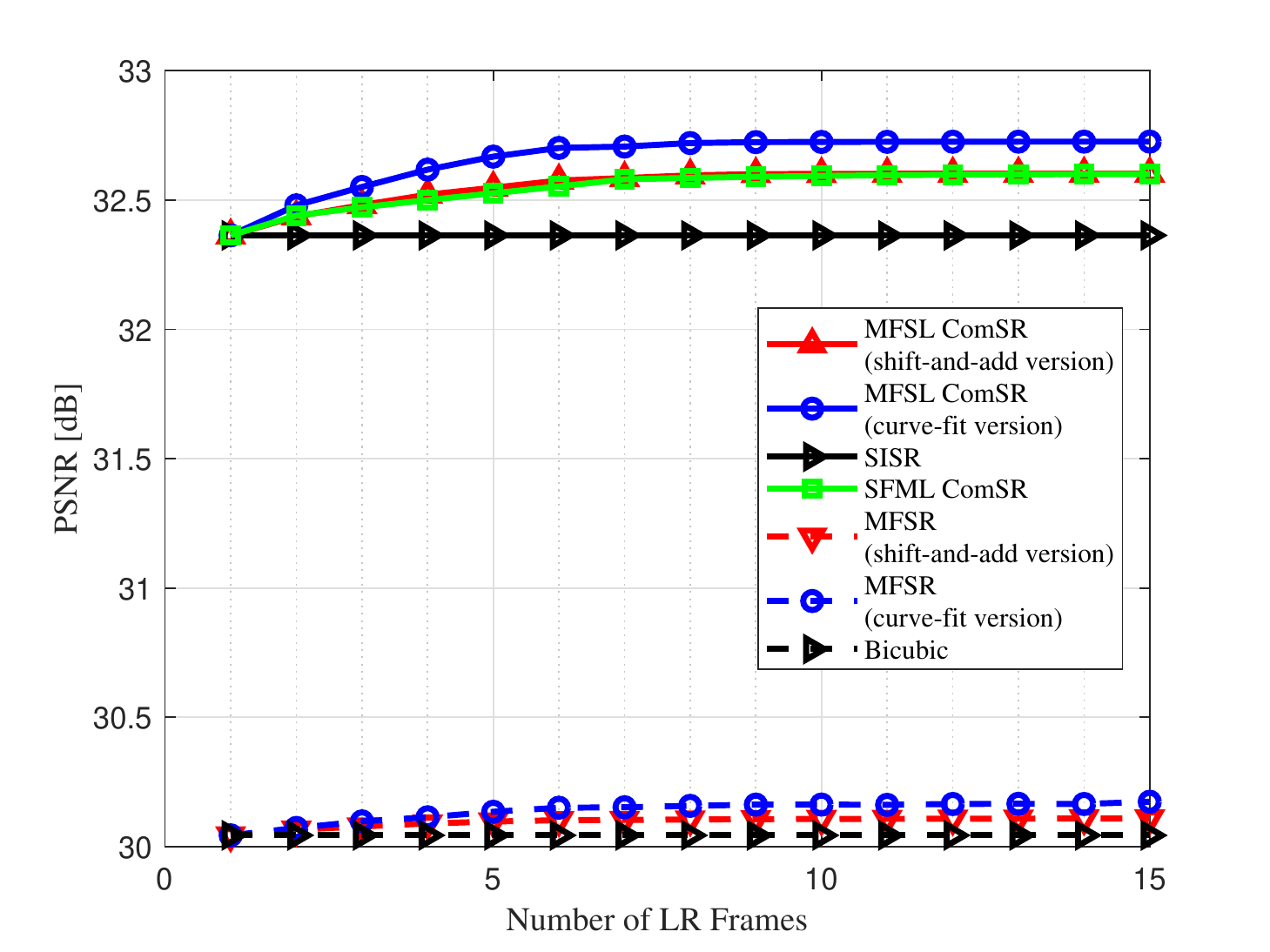}} & 
     		\subfloat[Set14, Noise-Free]{\includegraphics[width = 90 mm, scale = 1]{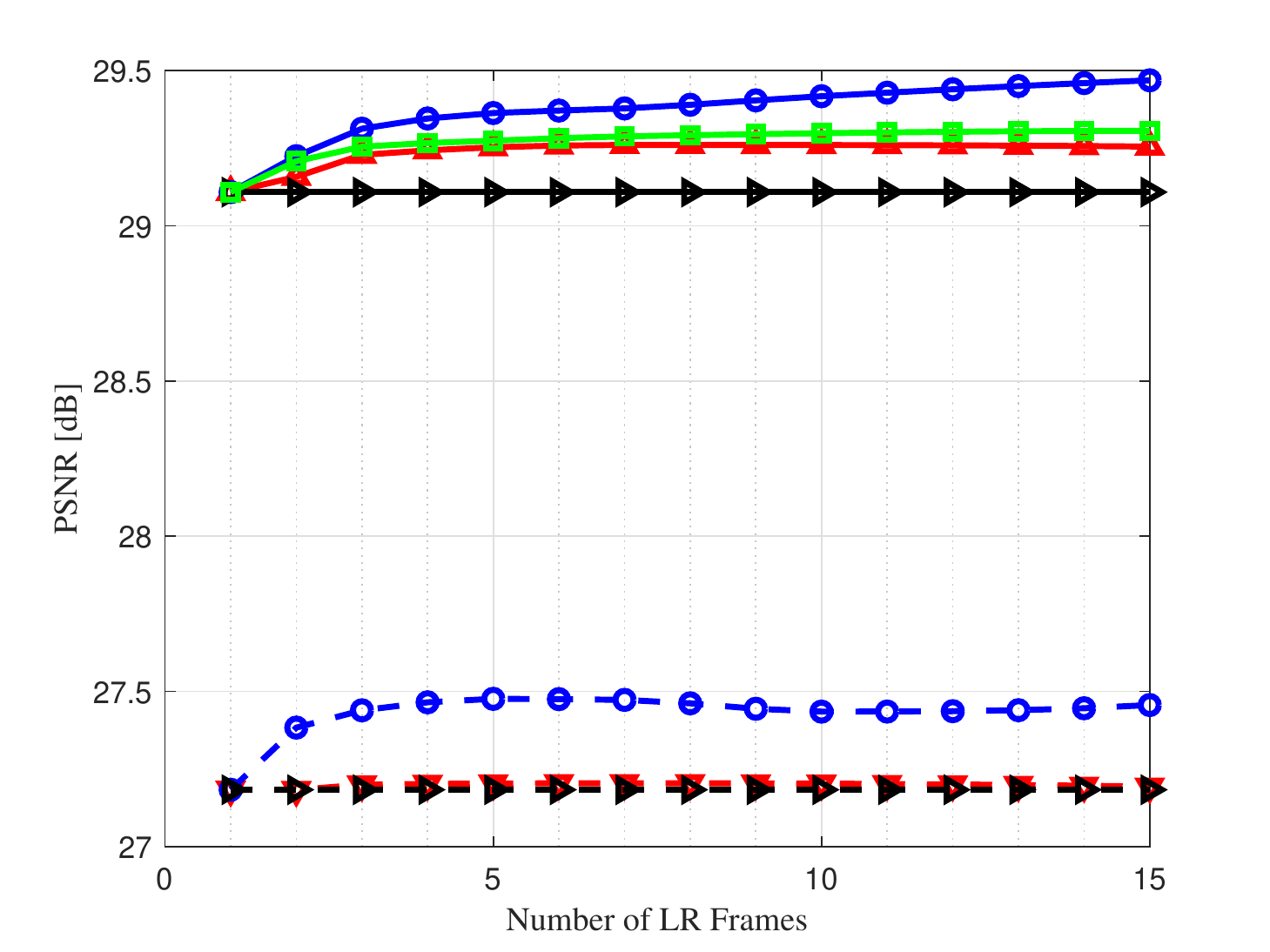}}  
     		\\
     		\subfloat[Set5, $\sigma_{n}=0.001$]{\includegraphics[width = 90 mm, scale = 1]{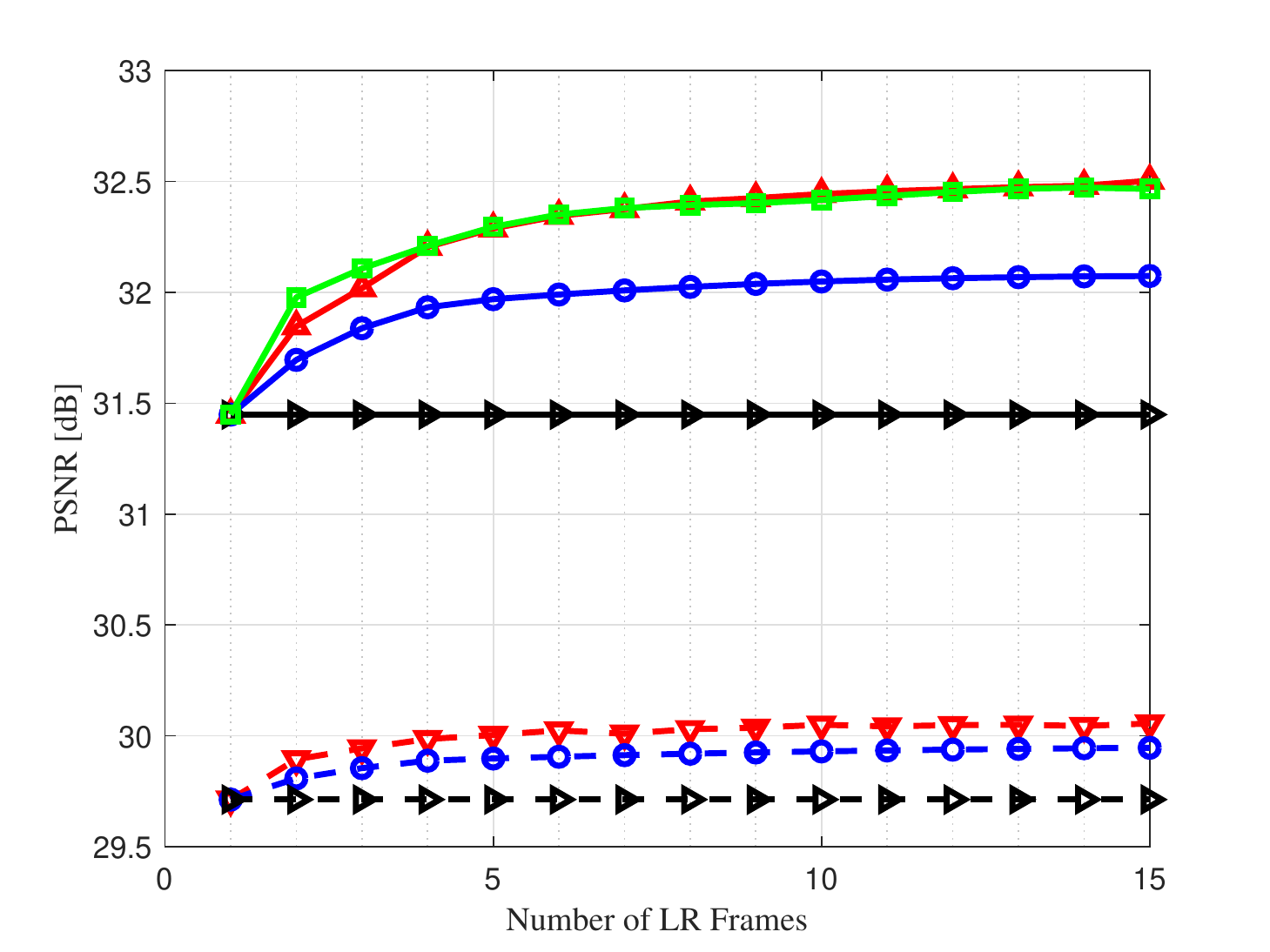}}  
     		& 
     		\subfloat[Set14, $\sigma_{n}=0.001$]{\includegraphics[width = 90 mm, scale = 1]{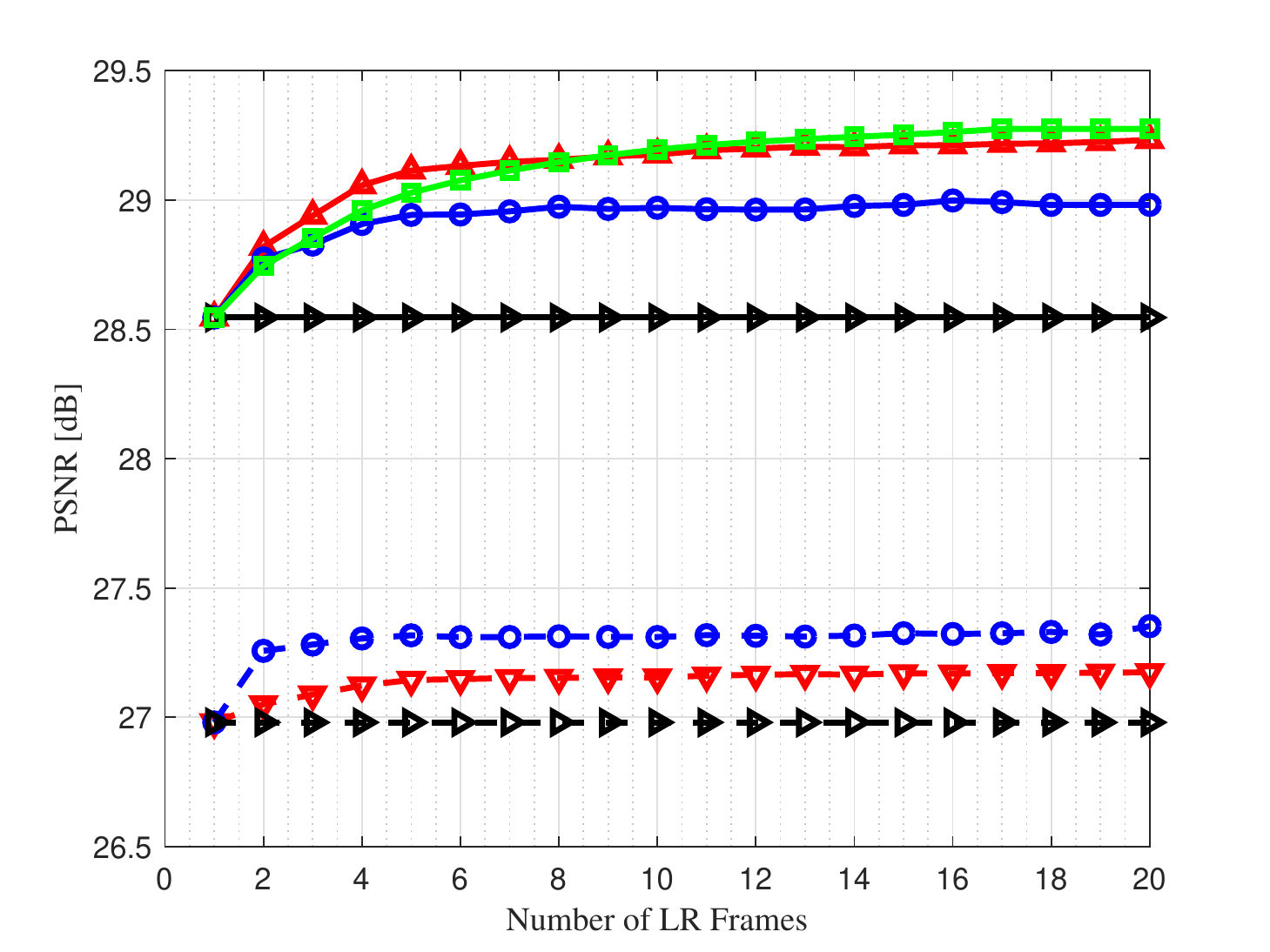}} 
     		\\
     		\subfloat[Set5, $\sigma_{n}=0.005$]{\includegraphics[width = 90 mm, scale = 1]{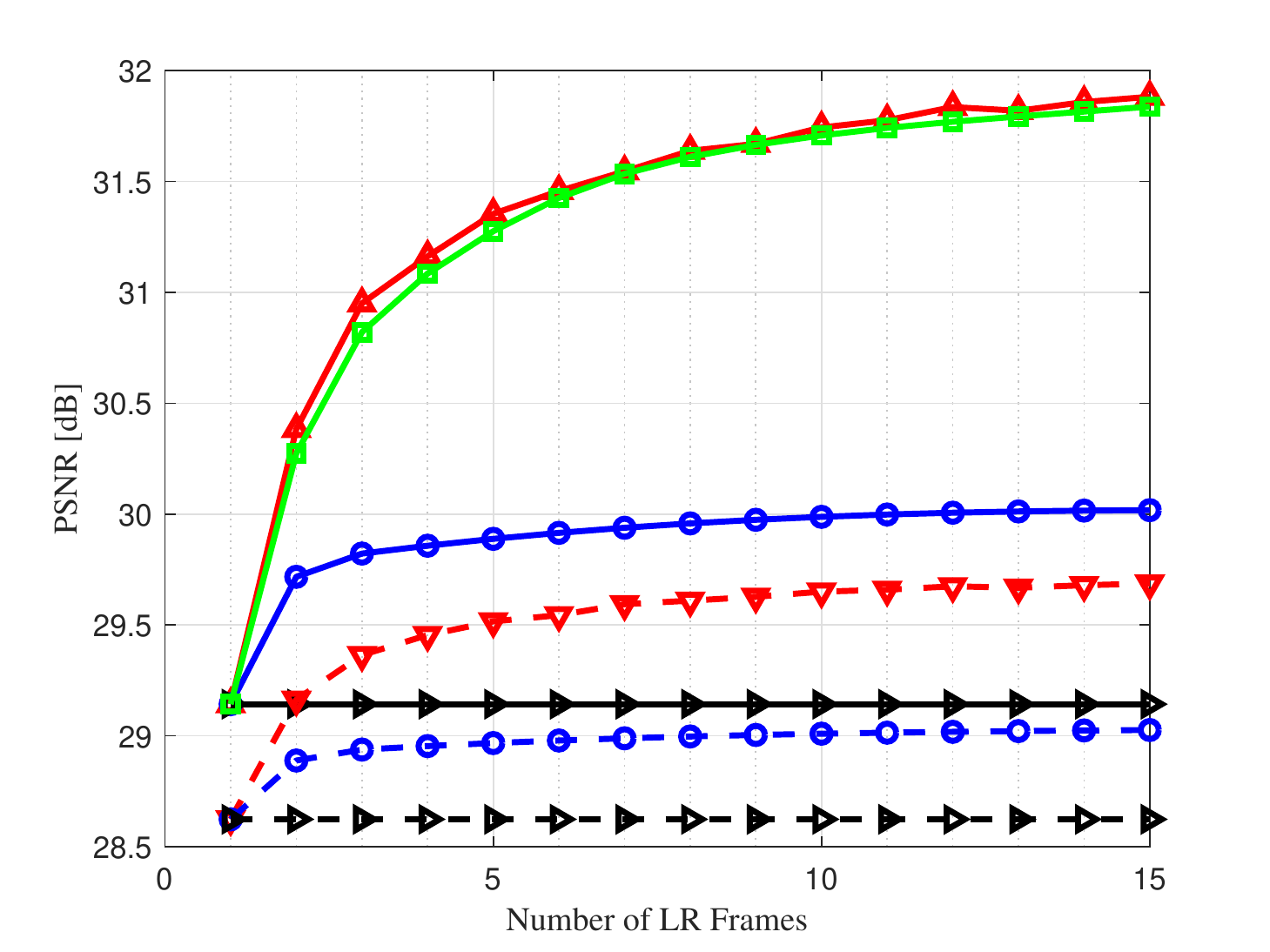}}  
     		& 
     		\subfloat[Set14, $\sigma_{n}=0.005$]{\includegraphics[width = 90 mm, scale = 1]{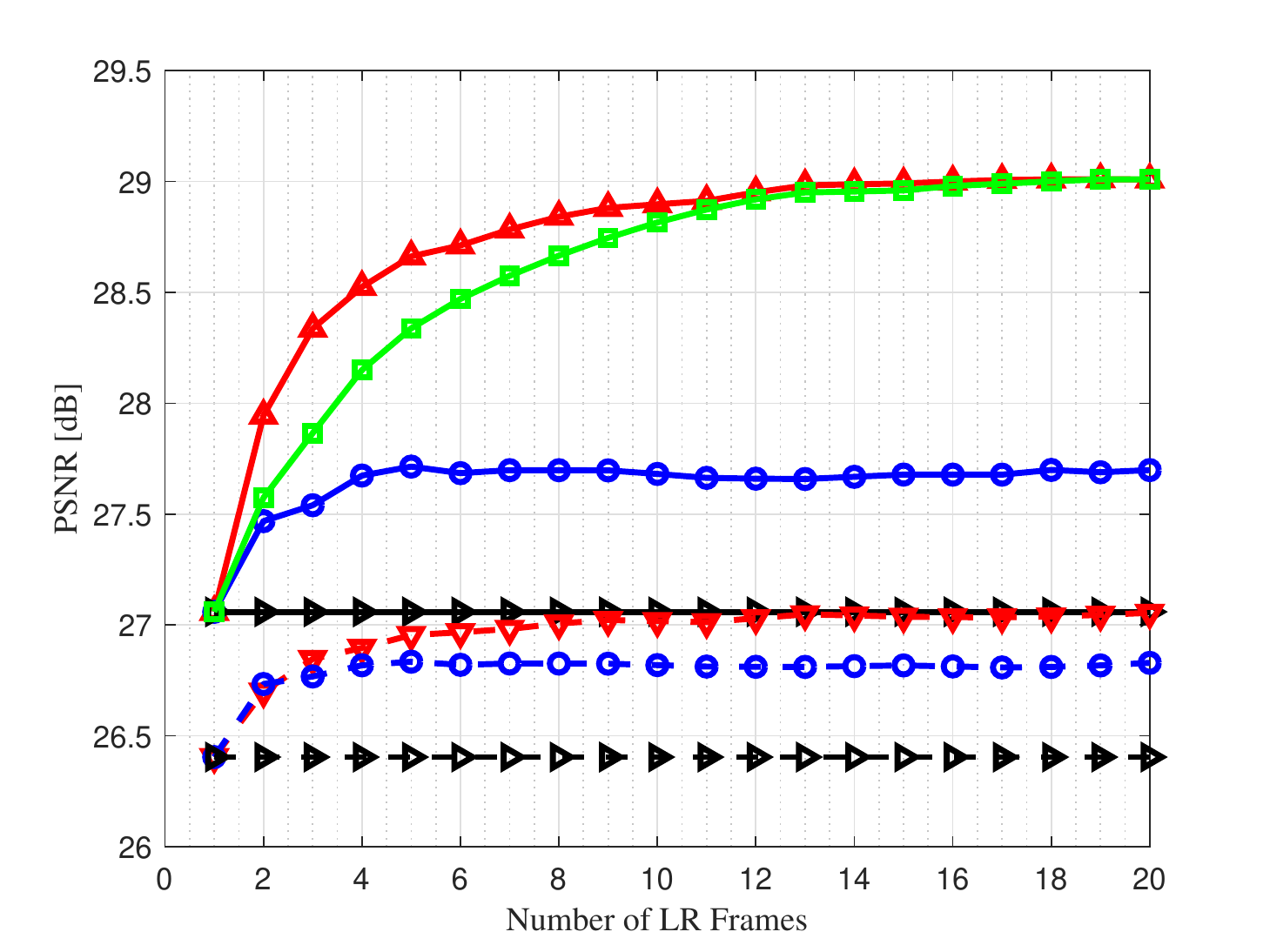}}
     	\end{tabular}
     	\caption{Average PSNR (in dB) vs number of LR frames for up-scaling factor $\times3$. Simulation results on Set5 dataset for noise-free, $\sigma_{n}=0.001$, and $\sigma_{n}=0.005$ are illustrated in (a), (c), and (e), respectively.
    Simulation results on Set14 dataset for noise-free, $\sigma_{n}=0.001$, and $\sigma_{n}=0.005$ are illustrated in (b), (d), and (f), respectively.}
     	
     	\label{sim_x3}
     \end{figure*}

We evaluate the performance of the optimal method (MFSL ComSR) in comparison with other existing ones, for up-scaling factors 2 and 3 in Fig.~\ref{sim_x2} and Fig.~\ref{sim_x3}, respectively. Each column in Fig.~\ref{sim_x2} or Fig.~\ref{sim_x3} corresponds to a dataset, and each row corresponds to a specific amount of noise in the LR images. The horizontal axis of each sub-graph shows the number of LR images, while the vertical axis demonstrates the average PSNR.   Following conclusions can be made based on Fig.~\ref{sim_x2} and Fig.~\ref{sim_x3}. 1) The more the number of LR images, the better the performance, but it has a saturation trend. 2) The general trend of the graphs is independent of the dataset. 3) The higher the noise of the LR images, the lower the final PSNR. 4) The superiority of the MFSL ComSR method over the others becomes more apparent when the noise increases. 5) Each mode of the MFSL ComSR methods works better in either noisy or noise-free case. 6) There is a relatively large gap between PSNRs of the methods that use SISR in the last stage of their methods and those that do not. This observation justifies the use of SISR after MFSR in MFSL ComSR.

\begin{figure*}[htbp]
	\vspace{-8 pt}
	\centering
	\begin{tabular}{cc}	
		\subfloat[Set5, $\sigma_{n}=0.001$]{\includegraphics[width = 90 mm, scale = 1]{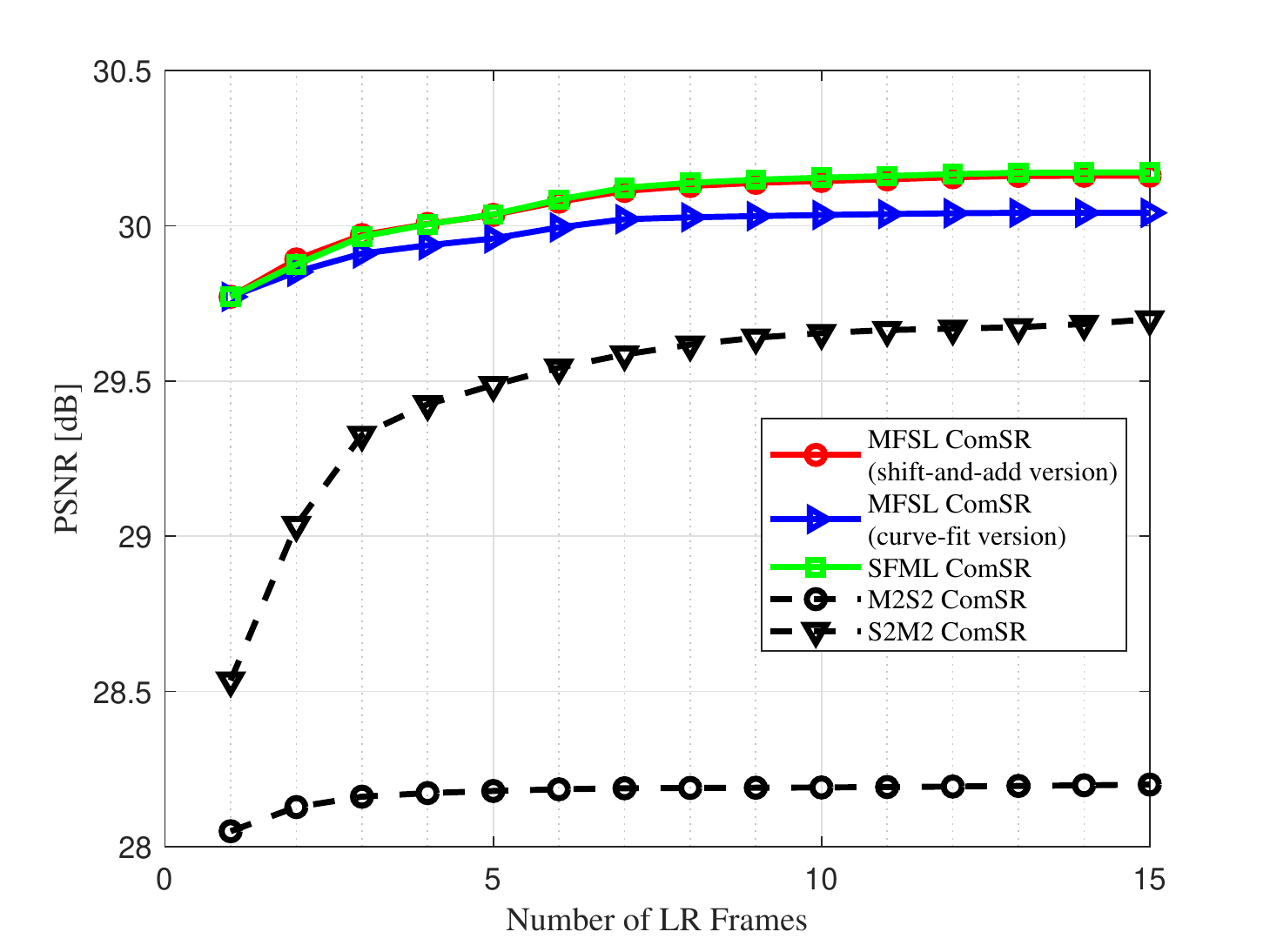}}  
		& 
		\subfloat[Set14, $\sigma_{n}=0.001$]{\includegraphics[width = 90 mm, scale = 1]{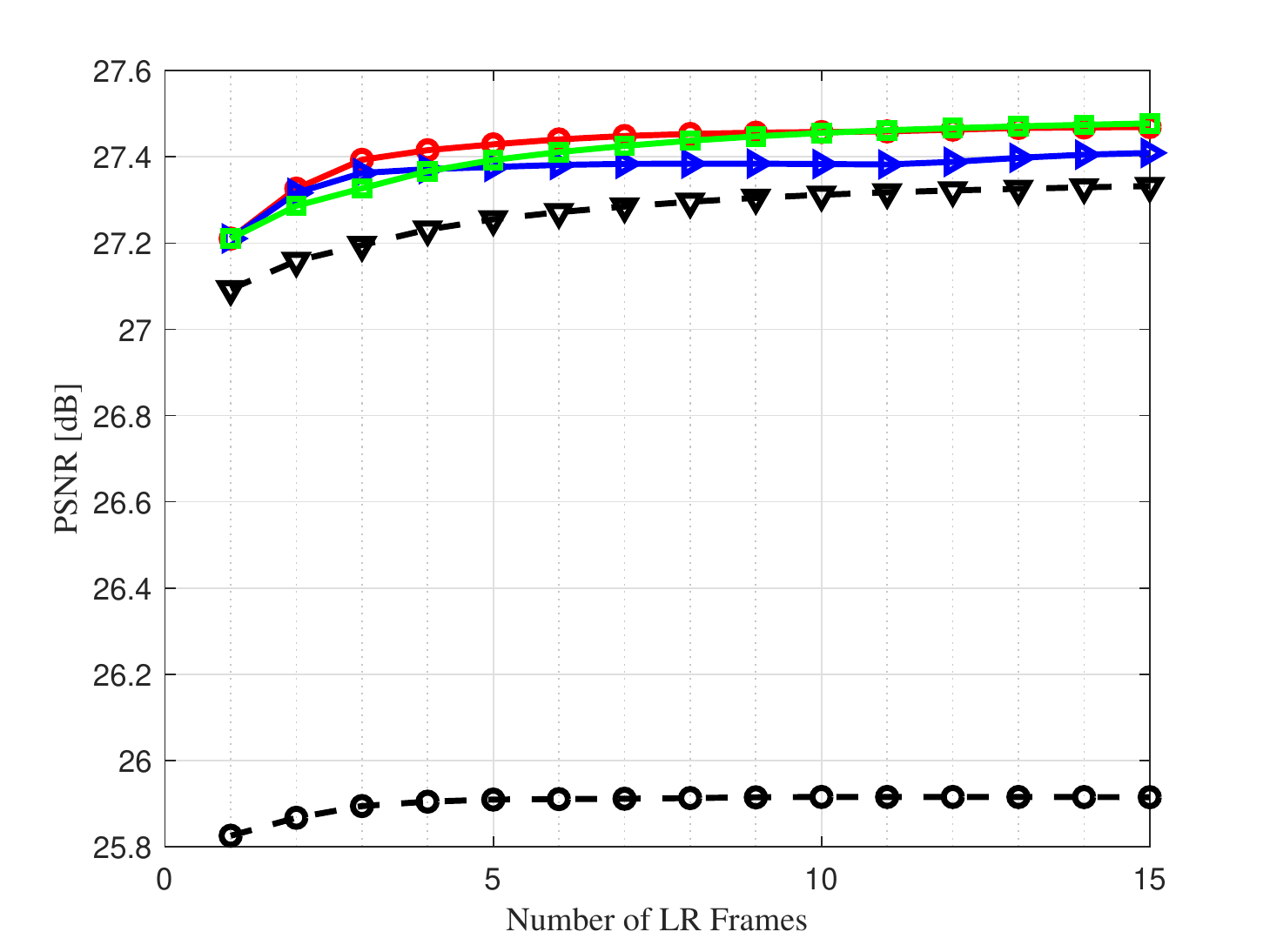}} 
		\\
		\subfloat[Set5, $\sigma_{n}=0.005$]{\includegraphics[width = 90 mm, scale = 1]{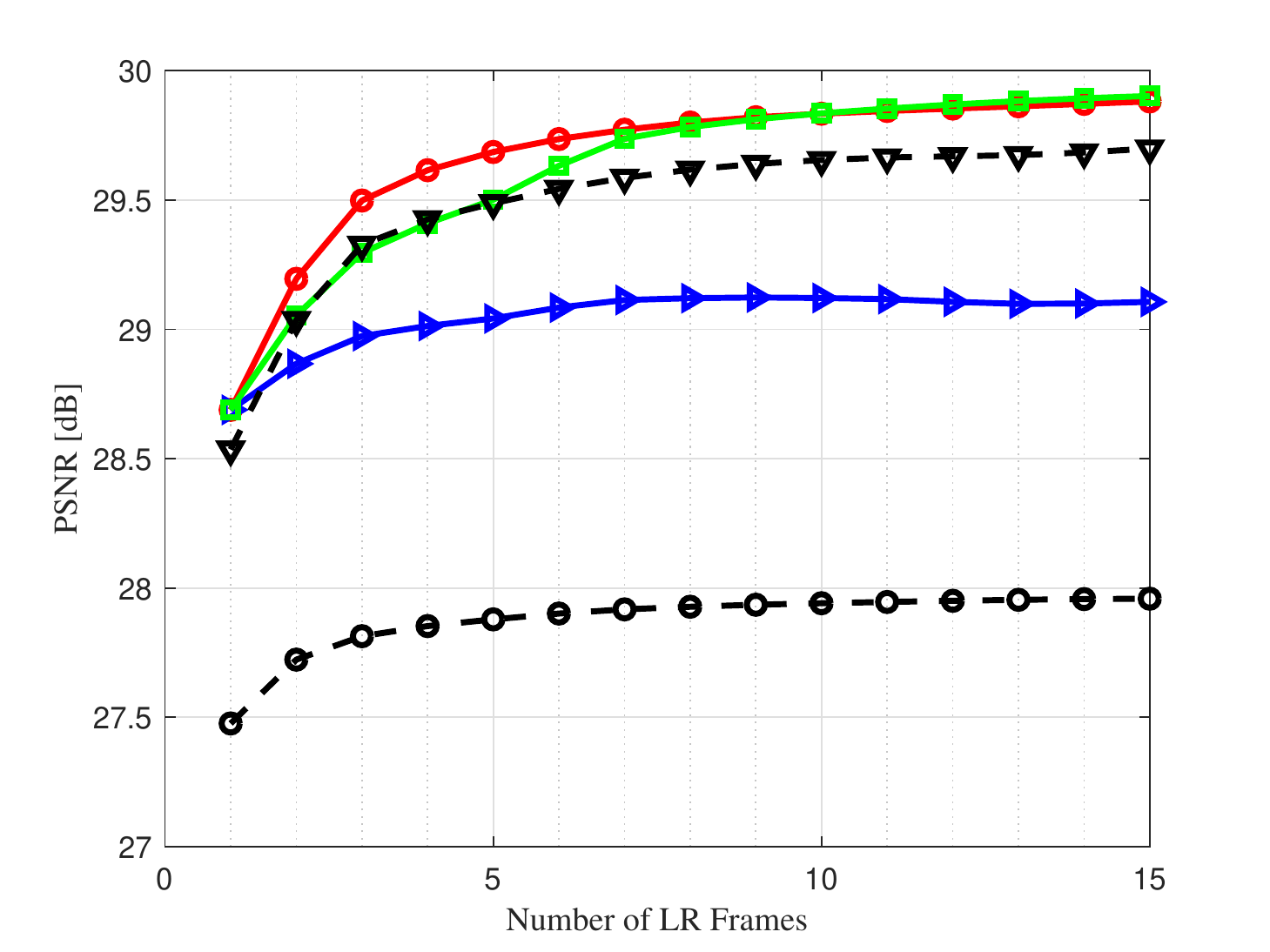}}  
		& 
		\subfloat[Set14, $\sigma_{n}=0.005$]{\includegraphics[width = 90 mm, scale = 1]{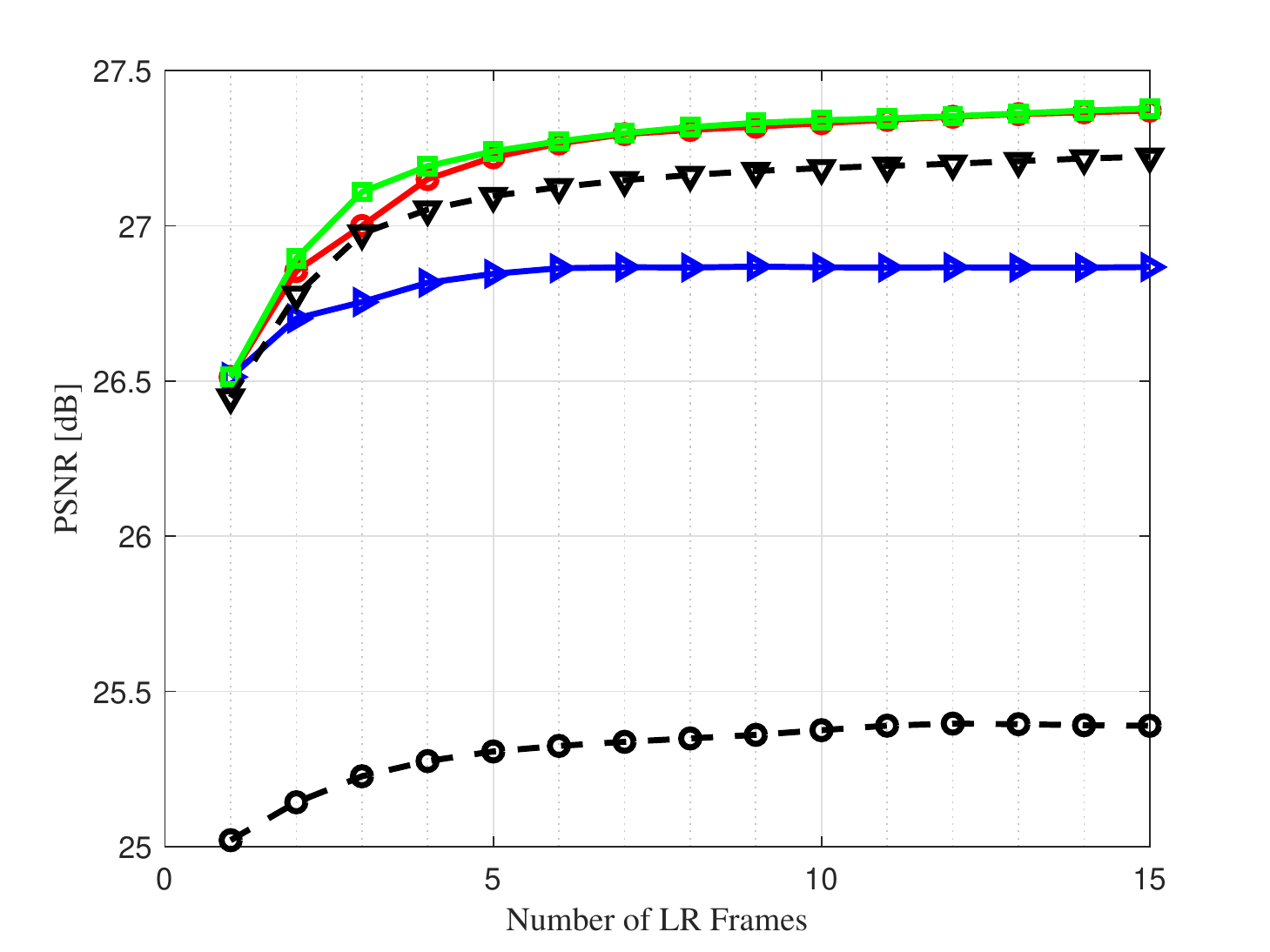}}
	\end{tabular}
	\caption{Average PSNR (in dB) vs number of LR frames for up-scaling factor $\times4$. Simulation results are on Set5 and Set14 datasets for two different amount of noises $\sigma_{n}=0.001$ and $\sigma_{n}=0.005$.}
	
	\label{sim_x4}
\end{figure*}

\subsection{Up-scaling factor is a divisible integer}

To examine the MFSL ComSR method thoroughly, we conduct some simulations when $r$ is divisible. Fig.~\ref{sim_x4} depicts the evaluation results of the MFSL ComSR method on different datasets with different amounts of noise. As it can be seen from Fig.~\ref{sim_x4}, MFSL ComSR has better performance than the other methods. SFML ComSR method is very close to MFSL ComSR in terms of PSNR, but it initially applies SISR to each LR image, so it has higher computational complexity. In the case of the up-scaling factor of 4, the slope of the PSNR curve significantly decreases for the large number of LR images, which is not the case in up-scaling factors 2 or 3.

\subsection{Simulations on real-world data}
\begin{figure*}[bp]
\vspace{-6 pt}	
	\centering
	\begin{tabular}{ccc}
		\centering
		\subfloat[SISR (SRCNN)]{\includegraphics[width= 5.5cm,height=4cm]{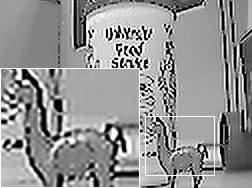}} 
		&
		\subfloat[MFSR (Curve-Fit)]{\includegraphics[width= 5.5cm,height=4cm]{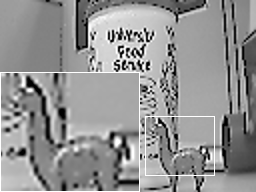}} 
		&
		\subfloat[MFSL ComSR (Curve-Fit version)]{\includegraphics[width= 5.5cm,height=4cm]{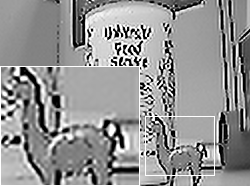}}  
  \\
		\subfloat[MFSR (Shift-and-Add)]{\includegraphics[width= 5.5cm,height=4cm]{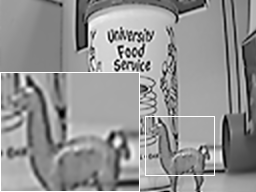}}  
		& 
		\subfloat[SFML ComSR]{\includegraphics[width= 5.5cm,height=4cm]{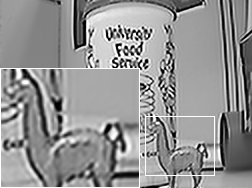}}
		&
		\subfloat[MFSL ComSR (Shift-and-Add version)]{\includegraphics[width= 5.5cm,height=4cm]{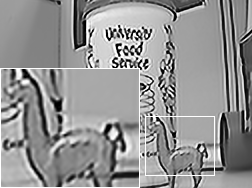}} 
		
	\end{tabular}
	\caption{Applying different SR methods on a real-world dataset (Alpaca \cite{MFSR_Farsiu_2004}) for up-scaling factor $\times2$.}
	
	\label{alpaca}
\end{figure*}

\begin{figure*}[htbp]
	\centering
	\begin{tabular}{ccc}
		\centering
		\vspace{-0 pt}	
		\subfloat[Bicubic]{\includegraphics[width= 5.5cm]{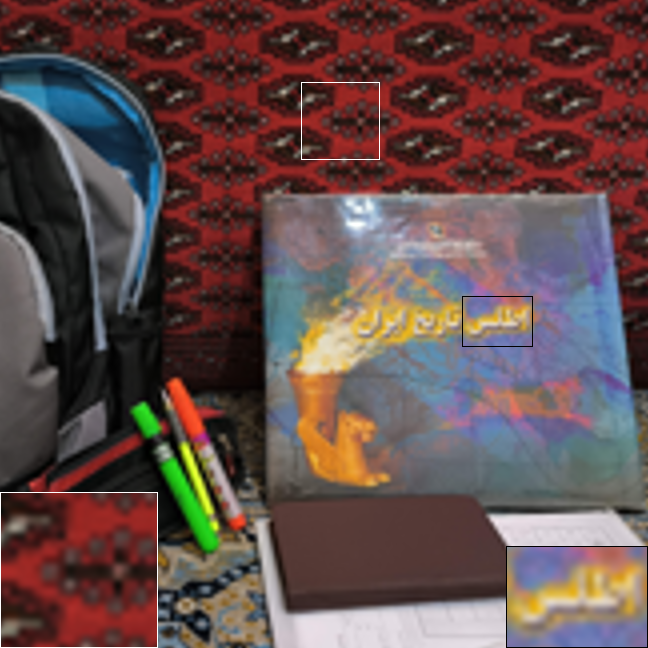}}  
		& 
		\subfloat[SISR]{\includegraphics[width= 5.5cm]{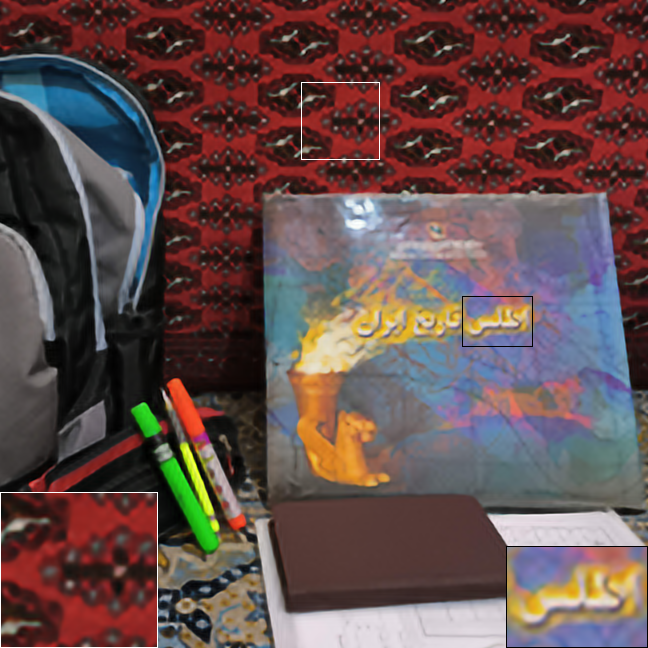}} 
		&
		\subfloat[MFSR (Curve-Fit)]{\includegraphics[width= 5.5cm]{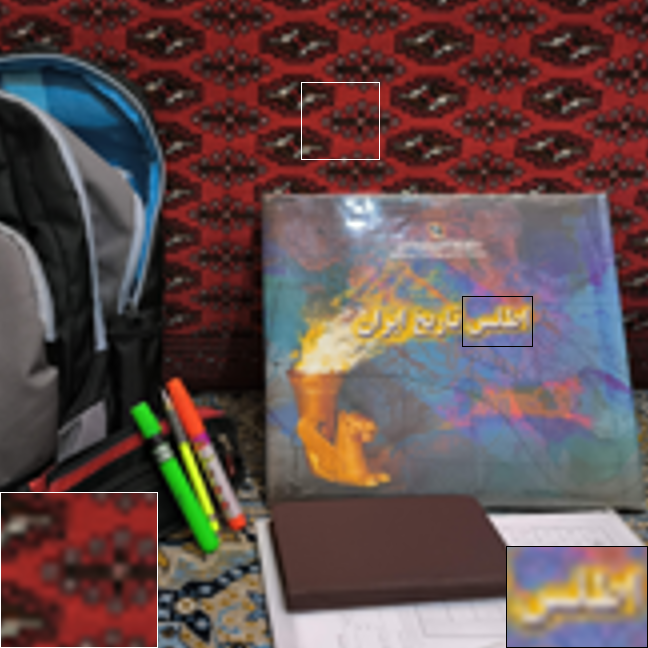}} 
		
		\\
		
		\subfloat[MFSR (Shift-and-Add)]{\includegraphics[width= 5.5cm]{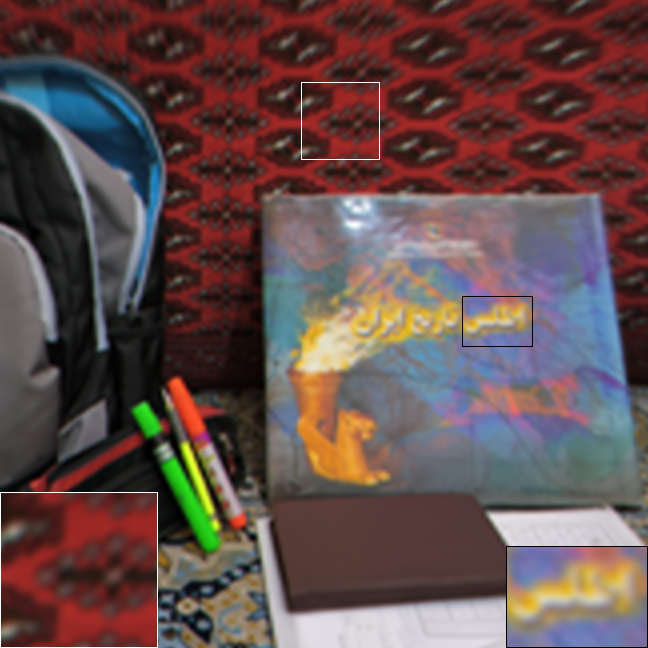}}

		&
		  \subfloat[S2M2 ComSR]{\includegraphics[width= 5.5cm]{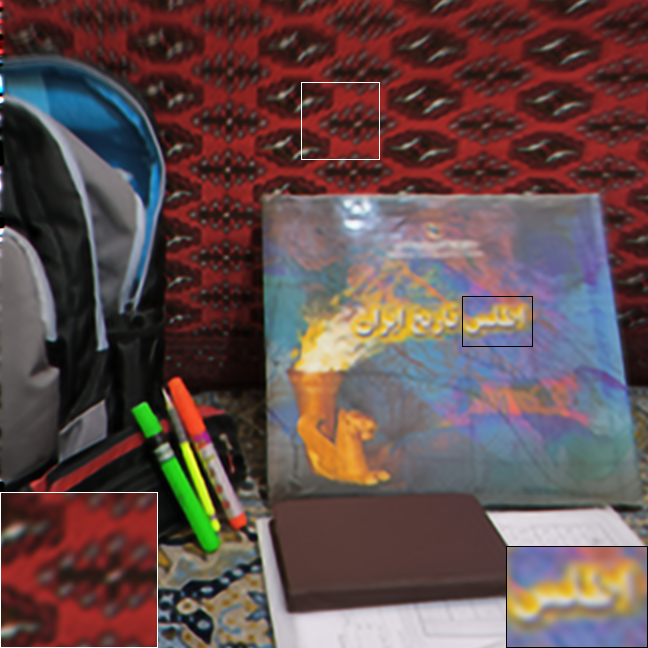}} 
		& 
		\subfloat[M2S2 ComSR]{\includegraphics[width= 5.5cm]{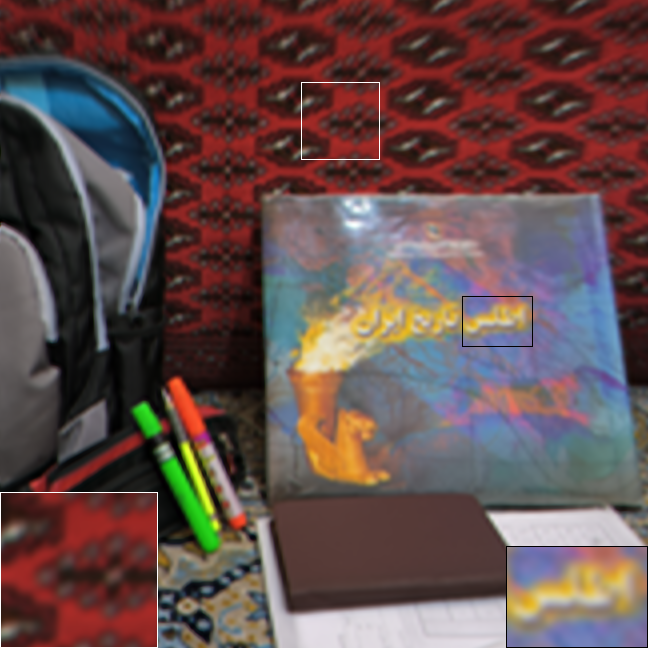}}
		
		\\
		
		\subfloat[SFML ComSR]{\includegraphics[width= 5.5cm]{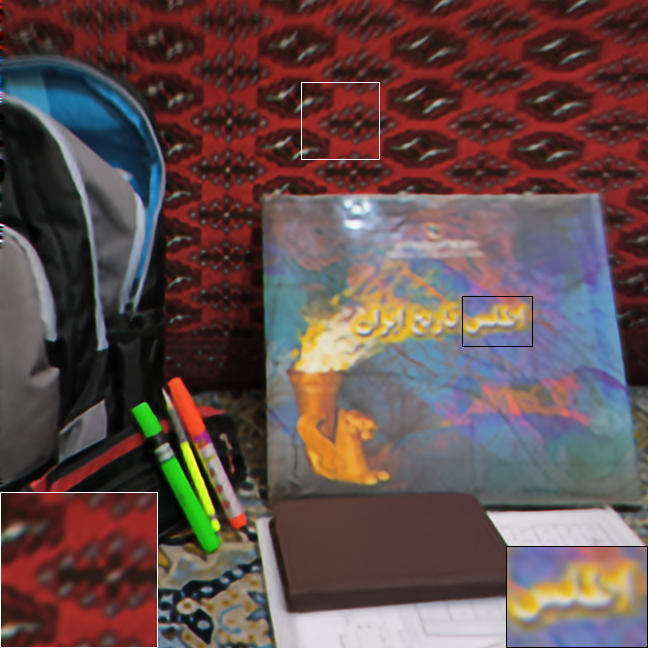}}
		&

        \subfloat[MFSL ComSR (Shift-and-Add version)]{\includegraphics[width= 5.5cm]{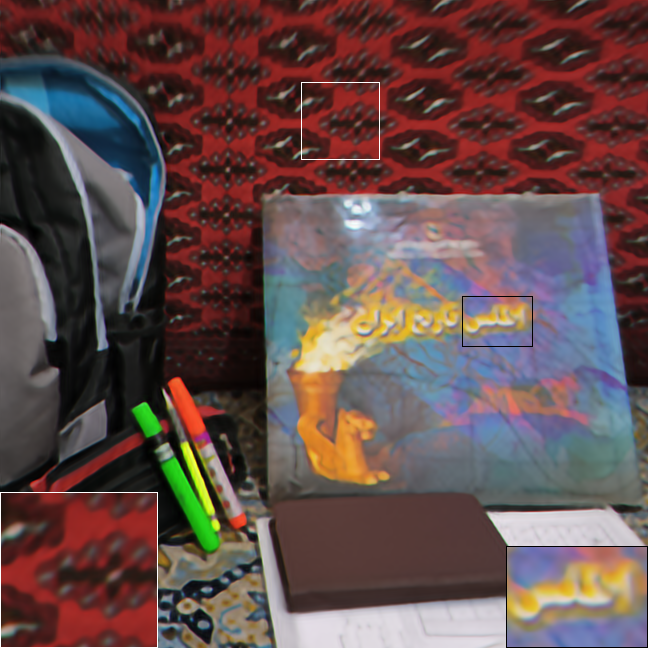}} 
		&
		\subfloat[MFSL ComSR (Curve-Fit version)]{\includegraphics[width= 5.5cm]{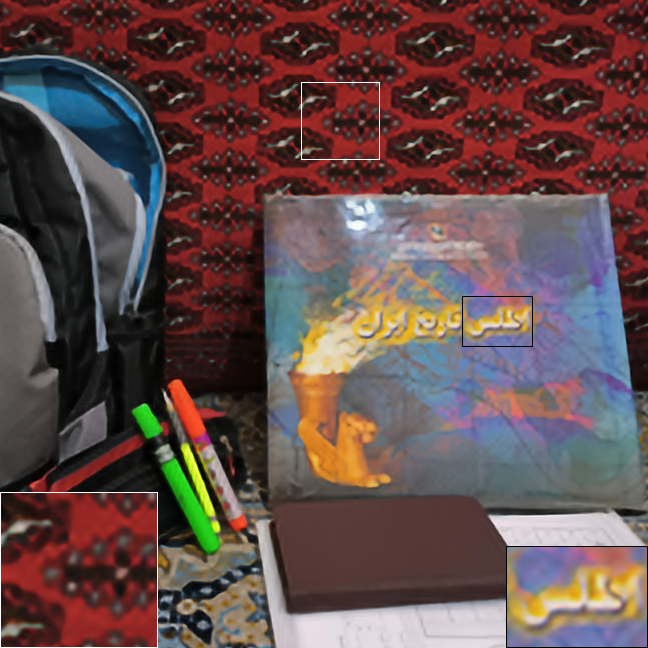}} 
		
	\end{tabular}
	\caption{Applying different SR methods on a real-world dataset (captured by burst shot) for up-scaling factor $\times4$.}
	
	\label{phone}
\end{figure*}

In this part, we test the MFSL ComSR method on the two real-world (non-synthetic) data. In Fig.~\ref{alpaca}, the noisy Alpaca dataset of \cite{MFSR_Farsiu_2004} (with 55 LR images) is used for the up-scaling factor of 2. Up-scaling factor 4 is investigated on our almost noise-free dataset (10 LR images captured by burst shot tool of Xiaomi POCO X4 GT smartphone) in Fig.~\ref{phone}. Qualitative comparison on these real data also validates the high performance of the MFSL ComSR method. It can also be observed that shift-and-add version of the MFSL ComSR method works better for the noisy Alpaca dataset, while curve-fit version of the MFSL ComSR works better for our almost noise-free one.
 
\section{Conclusion}
 In this paper, we presented how to optimally combine SISR and MFSR based on an analytical perspective. Our mathematical analysis and simulation results showed that the best performance is obtained by MFSL ComSR, the method that combines LR images first (MFSR) and applies SISR to the resulting image. This approach also has significantly lower computational complexity than the methods that first apply SISR to all the input LR images. We proposed two modes for MFSL ComSR, one for noisy images and the other for noise-free cases.


%

%


\ifCLASSOPTIONcaptionsoff
  \newpage
\fi

\newpage



\bibliographystyle{IEEEtran} 
%



%

\begin{IEEEbiography}[{\includegraphics[width=2in,height=2.5in,trim=4 4 4 4,clip,keepaspectratio]{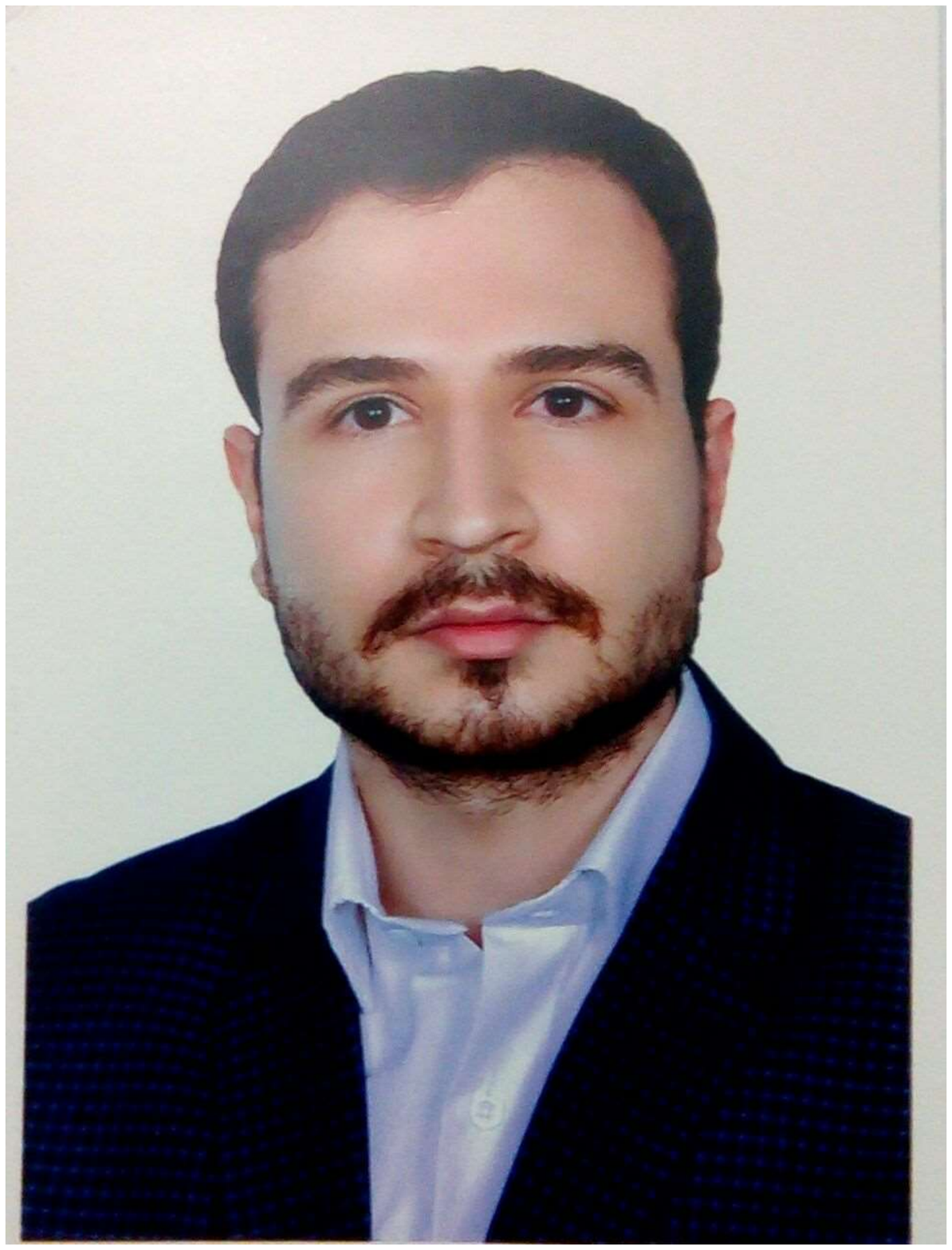}}]{Mohammad Mahdi Afrasiabi} Mohammad Mahdi Afrasiabi received the B.Sc. and M.Sc. degrees in electrical engineering (telecommunication) from University of Tehran, Iran in 2013 and 2015, respectively. He is working currently as a PhD candidate at University of Tehran in the field of telecommunication engineering. His main research interests include analytical approaches for solving signal and image processing problems, sparse coding, image super-resolution, image enhancement, and applications of machine learning in wireless communications.
\end{IEEEbiography}	

\begin{IEEEbiography}[{\includegraphics[width=1in, height=2.5in, trim=4 4 4 4,clip,keepaspectratio]{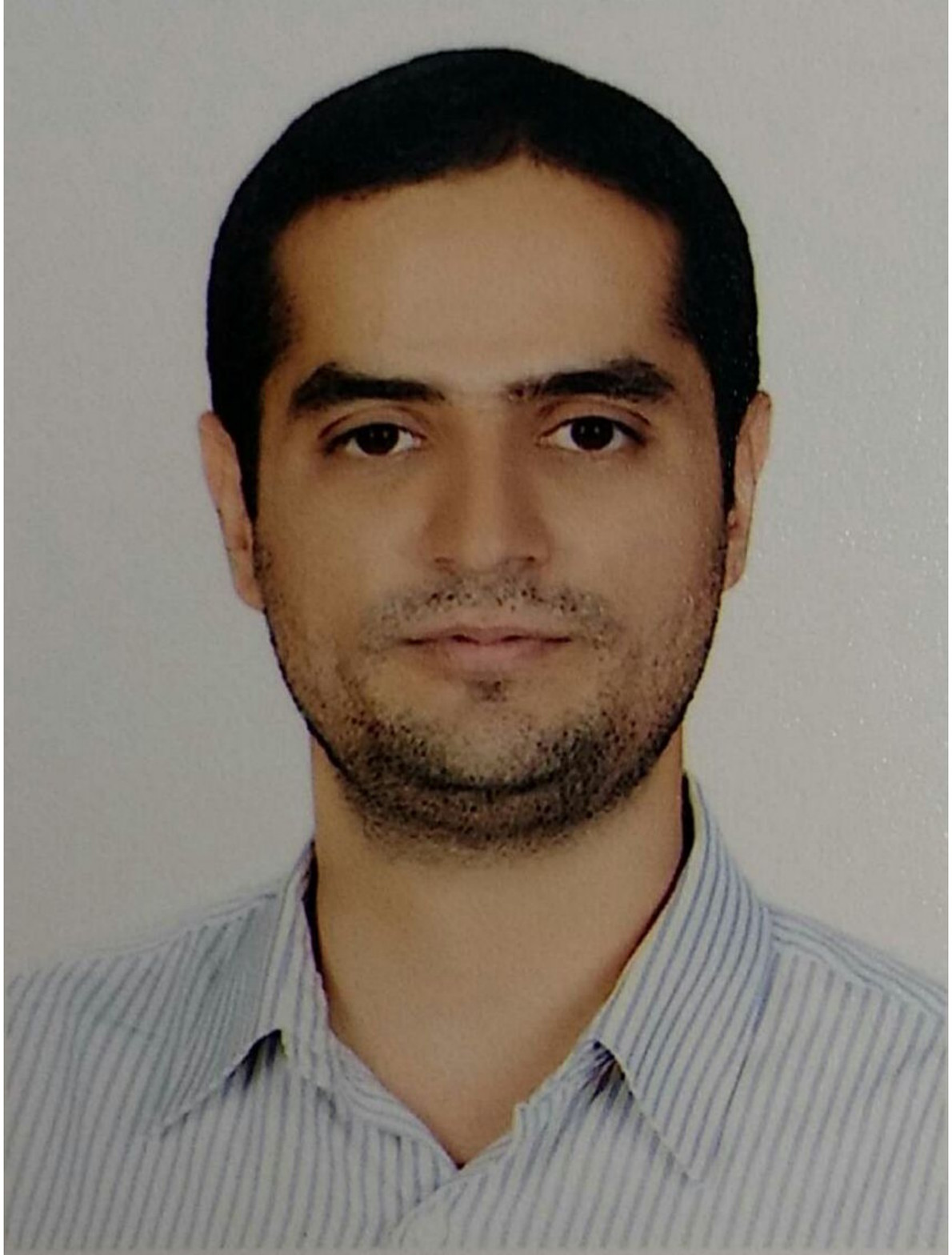}}]
{Reshad Hosseini}
Reshad Hosseini received the B.Sc. degree in electrical engineering (telecommunication) from the University of Tehran, Tehran, Iran, in 2004, and the Ph.D. degree from the Faculty of Electrical Engineering and Computer Science, Technical University of Berlin, Berlin, Germany, in 2012. He is currently an Assistant Professor with the School of Electrical and Computer Engineering, College of Engineering, University of Tehran. His professional interest topics are machine learning, signal processing, and computational vision. He is particularly interested in the mathematical foundation of these fields, such as differential geometry, optimization, functional analysis, and statistics. His current research interests include manifold optimization, large-scale mixture models, 3-D reconstruction, neural system identification, visual recognition using deep learning, and accelerating reinforcement learning.
\end{IEEEbiography}

\begin{IEEEbiography}
[{\includegraphics[width=1in, height=2.5in, trim=4 4 4 4,clip,keepaspectratio]{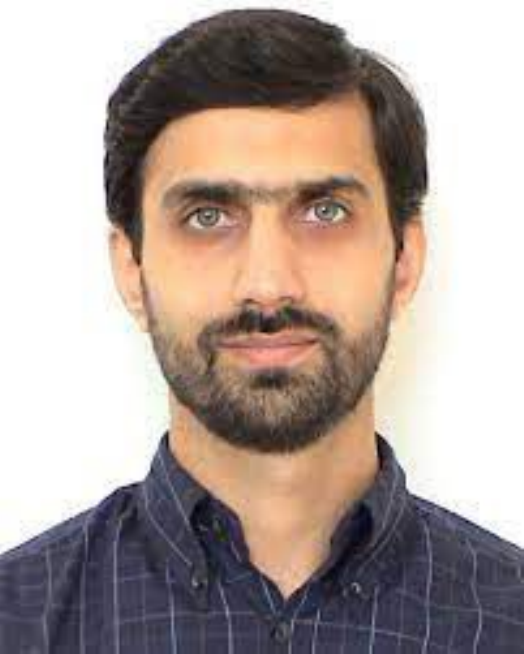}}]
{Aliazam Abbasfar} Aliazam Abbasfar (Senior Member, IEEE) received the B.Sc. (Highest Hons.) and M.Sc. degrees in electrical engineering from the University of Tehran, Tehran, Iran, in 1992 and 1995, respectively, and the Ph.D. degree in electrical engineering from the University of California at Los Angeles (UCLA), Los Angeles, CA, USA, in 2005.,From 2001 to 2004, he held positions as a Senior Design Engineer in the areas of communication system design and digital design with various startup companies in California. Upon graduation from UCLA, he joined Rambus Inc., Sunnyvale, CA, USA, where he was a Principal Engineer working on high-speed data communications on wireline serial and parallel links. He is currently an Associate Professor with the University of Tehran. 
His main research interests include wireless and wireline communications, error correcting codes, and VLSI for digital data communications.
\end{IEEEbiography}





\end{document}